\documentclass{article}
\usepackage[utf8]{inputenc}
\usepackage[parfill]{parskip}

\usepackage{amsmath,amsfonts,amsthm,fullpage,amssymb}
\usepackage{algorithm,amsmath,amssymb}
\usepackage{algorithmic}
\usepackage{graphicx, color}
\usepackage{url}
\usepackage[utf8]{inputenc}
\usepackage{amsmath, amsthm, amssymb, amsfonts}
\usepackage{bbm}
\usepackage{graphicx}
\usepackage{tabularx}
\usepackage{float}
\usepackage{subfigure}
\usepackage[parfill]{parskip}

\newpage

\title{Capturing the Flow of Art History}
\author{Chenxi Ji}
\date{December 4, 2022}

\begin{document}

\maketitle

\section{Introduction}

Do we really understand how machine classifies art styles? Historically, art is perceived and interpreted by human eyes and there are always controversial discussions over how people identify and understand art. Historians and general public tend to interpret the subject matter of art through the context of history and social factors. Style, however, is different from subject matter. Given the fact that Style does not correspond to the existence of certain objects in the painting and is mainly related to the form and can be correlated with features at different levels.(Ahmed Elgammal et al. 2018), which makes the identification and classification of the characteristics artwork's style and the ``transition"  -  how it flows and evolves - remains as a challenge for both human and machine.

In this project, a series of state-of-art neural networks and manifold learning algorithms are explored to unveil this intriguing topic: How does machine capture and interpret the flow of Art History?

\section{Data Preprocessing}

\subsection{Data Collection}

Data preprocessing plays an important part in this project. The process of collecting the source data is inspired by a Kaggle competition, which referred to Web Gallary of Art (https://www.wga.hu) and collected necessary Artwork data from public available website through simple web scraping scripts. 

To perform planned implementation in this project, two parts of data need to be collected. The first part is the set of image data of all the artworks to be explored. The second part is the annotation data summarizing the information of each artwork - the ``metadata" and potential ``labels".

Collecting annotation in this domain is hard since it requires expert annotators and typical crowd sourcing annotation are usually not qualified.(Ahmed Elgammal et al. 2018). So this annotation file is a pre-cleaned file available on Kaggle that can be directly use after basic preprocessing. 

\subsection{Data Quality Check}

The raw dataset consists of a collection of 3198 data points,  each represents an artwork with the dimension of file size in pixels. The image data are stored as array-like objects to satisfy the input data requests for the neural network models to be explored in later steps. 

The standard input format of neural network models for image data is $224 \times 224$, however, not every artwork follow this size and aspect ratio. During the initial data reprocessing stage, there are some ``bad data" that could not fit the dimension/size requirement of the pre-defined  neural network model input hence were excluded from the input tensor.

The annotation data was directly collected from Kaggle with basic data cleaning and preprocessing, then loaded as a dataframe. The annotation data provides the following information describing the art work:

\begin{tabular}{|c|c|}
\hline
\textbf{Data} & \textbf{Example}  \\\hline

unique key & 1034 \\\hline
artist name &  ANGELICO, Fra \\\hline
artwork title &  madonna of the star \\\hline
picture summary data  & c. 1424, tempera and gold on panel, 84 x 51 cm, museo di san marco, florence \\\hline
file info & 803*1400, true color, 191 kb \\ \hline
link to the file in public website & https://www.wga.hu/art/a/angelico/12/00\_madon.jpg \\\hline
life marks of artist & (c. 1400-1455) \\\hline
active period of the artist & Early Renaissance \\\hline
school of the artist & painter \\\hline
active geographic location & Florence \\ \hline
nationality of the artist & Italian \\\hline

\end{tabular}

To summarize the annotation data, the dataset focuses specifically in European art history and contains the artworks from the period listed below. The number of artworks in each period are ranging from 300 to 700 per category.  

\begin{tabular}{|c|c|c|}
\hline
\textbf{Name} & \textbf{Time Period} & \textbf{Number of artworks} \\\hline
    Medieval & $12^{th}$ century & 721\\\hline
    Early Renaissance & $146^{th}-16^{th}$ century & 448\\\hline
    Northern Renaissance & $14^{th}-16^{th}$ century & 385\\\hline
    Baroque & $17^{th}$ century & 724\\\hline
    Romanticism & $19^{th}$ century & 302\\\hline
    Impressionism & $19^{th}$ century & 618	\\\hline
\end{tabular}

It is not surprising that some artworks from adjacent eras have similar styles that are relatively challenging to identify (for example, Early Renaissance and Northern Renaissance) hence it is necessary to perform certain merges so that each category have a relatively distinct art style. In Elgammal et al. (2018), post-impressionism and Pointillism are merged as post-impressionism; Cubismm, Analytical Cubism and Synthetic Cubism are merged as Cubismn; Realism, Contemporary Realism and New Realism are merged as Realism. Abstract Expressionism and Action Painting are merged as Abstract-Expressionism. Given the dataset used in this project is relatively small, Early Renaissance and Northern Renaissance will be considered for merging in next steps. 

From the artist perspective, the dataset studies the artwork from a total of 12 artists with 6 nationalities, Each artist has 100-500 artworks. 

\begin{tabular}{|c|c|c|c|}
\hline
\textbf{Name} & \textbf{Category} & \textbf{Nationality}  & \textbf{Number of artworks} \\\hline
    Fra Angelico & Early Renaissance & Italian & 244\\\hline
    Hieronymus Bosch & Northern Renaissance & Dutch & 162\\\hline
    Sandro Botticelli & Early Renaissance & Italian & 204\\\hline
    Pieter Bruegel the Elder & Northern Renaissance	 & Belgian & 223 \\\hline
    Flemish, Caravaggio & Baroque & Italian & 185\\\hline
    Eugène Delacroix & Romanticism & French & 105\\\hline
    Duccio di Buoninsegna & Medieval & Italian & 170\\\hline   
    Giotto di Bondone & Medieval & Italian & 551\\\hline 
    Francisco José de Goya y Lucientes & Romanticism & Spanish & 420\\\hline    
    Claude Monet & Impressionism & French & 198\\\hline  
    Vincent Van Gogh & Impressionism & Dutch & 420 \\ \hline
    Rembrandt Harmenszoon van Rijn & Baroque & Dutch & 539\\ \hline
\end{tabular}

\section{Methodology}

Convolutional neural network models and their variants have played a revolutionary role in advancing artificial intelligence.(A. Elgammal et al. 2018). 

While the Transformer architecture has become the de-facto standard in Natural Language processing tasks, it also shows great performance on image classification tasks and could attain excellent results compared to convolutional networks with substantially fewer computational resources. (A. Dosovitskiy et al., 2021) 

The main approach in is project is to choose a range of state-of-art convolutional deep neural network and transformers and perform art style identification and classification of the source data. Then visualize through various manifold learning algorithms to see if we can observe relationships between the artworks and if the machines can observe the flow of art history.

\subsection{Neural Network - Overview}

\textbf{AlexNet}

\begin{figure}[!ht]
    \centering
    \includegraphics[width=0.4\textwidth]{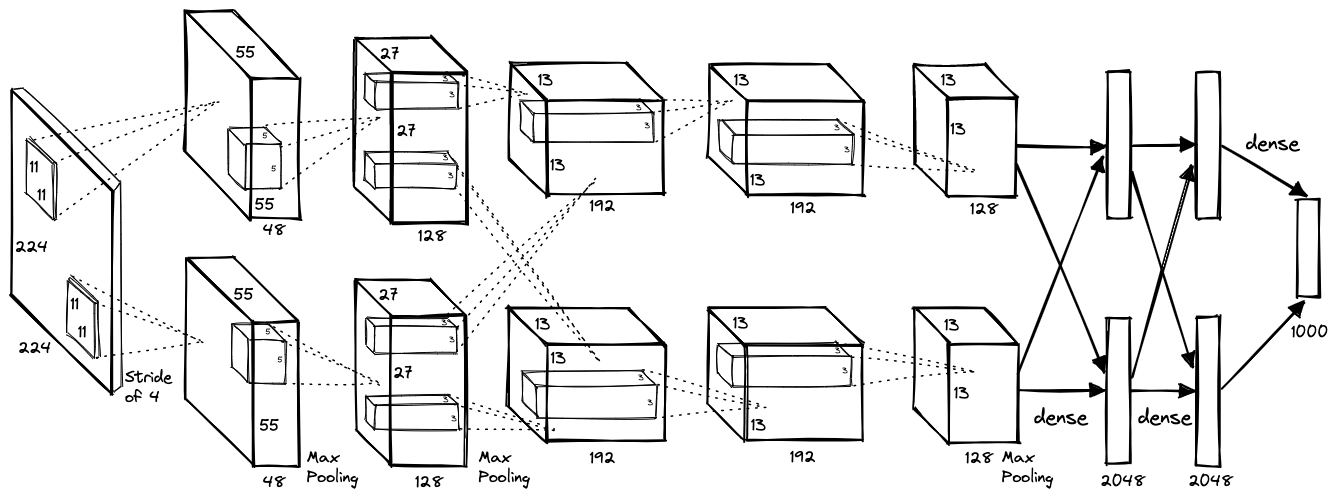}
    \caption{AlexNet Structure}
    \label{fig:alexnet_structure}
\end{figure}

AlexNet is a large, deep convolutional that classifies 1.2 million high-resolution images in the ImageNet LSVRC-2010 contest into the 1000 different classes. The structure is shown in  \ref{fig:alexnet_structure}, this neural network has 60 million parameters and 650,000 neurons and consists of 5 convolutional layers (A. Krizhevsky et al., 2012). 
              
\textbf{Resnet}

Comprehensive empirical evidence showed that residual neural networks are easier to optimize and can gain accuracy from considerably increased depth with relatively low complexity. (K. He et al., 2015).

\begin{figure}[!ht]
    \centering
    \includegraphics[width=0.4\textwidth]{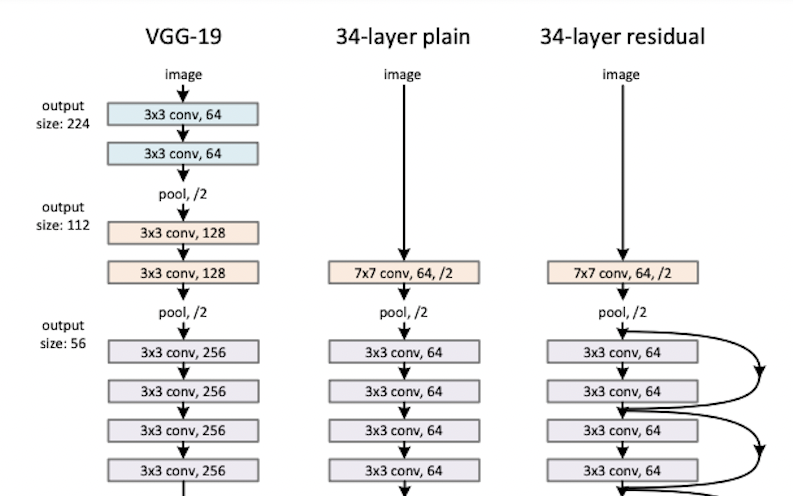}
    \caption{Resnet Structure (Partial)}
    \label{fig:resnet_partial}
\end{figure}

\begin{figure}[!ht]
    \centering
    \includegraphics[width=0.5\textwidth]{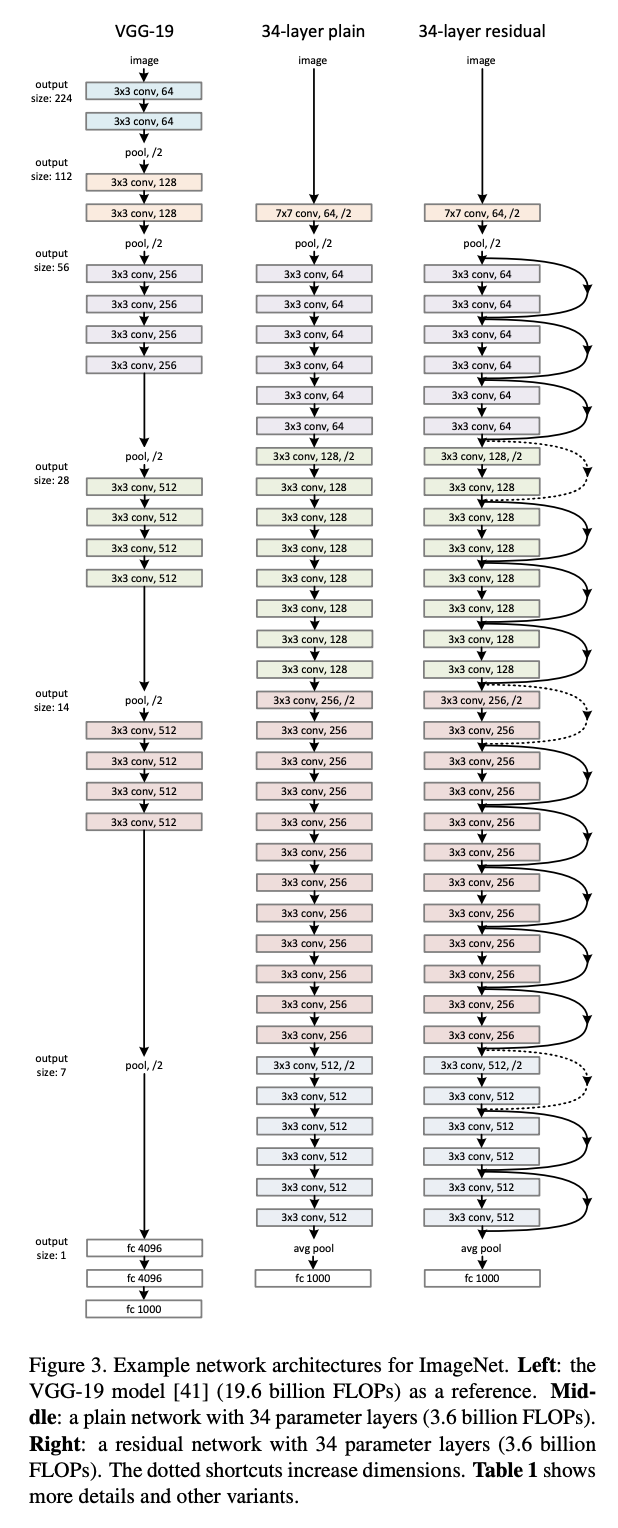}
    \caption{Resnet Structure}
    \label{fig:resnet_structure}
\end{figure}

As the structured shown in Figure \ref{fig:resnet_structure} and partially shown in Figure \ref{fig:resnet_partial}  the structures of Resnet are mainly inspired by the philosophy of VGG nets, yet with fewer filters and lower complexity.(K. He et al., 2015)

\textbf{VisionTransformer}

\begin{figure}[!ht]
    \centering
    \includegraphics[width=0.4\textwidth]{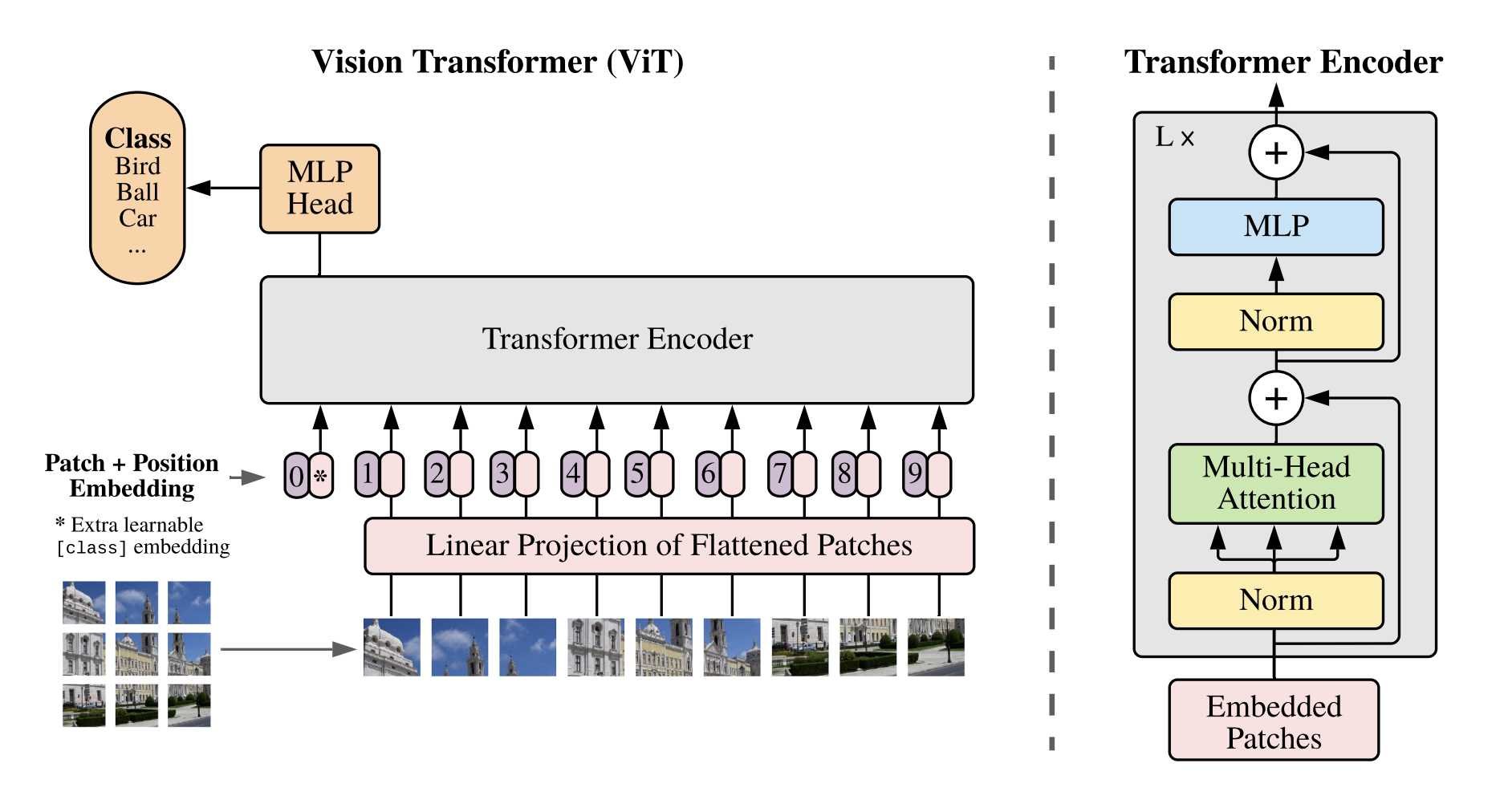}
    \caption{Vision Transformer Structure}
    \label{fig:VIT}
\end{figure}

Vision Transformer leverages a similar approach as Transformers in NLP to analyze images by splitting them into patches just like the tokens in NLP sentences. Then vision transformer provides the sequence of linear embeddings of these patches as an input. (A. Dosovitskiy et al., 2020). The structure of vision transformer is shown in Figure \ref{fig:VIT}.

\subsection{Manifold Learning Algorithms - Overview}

\textbf{Spectral Embedding}

Spectral Embedding is a non-linear dimensionality reduction method which forms an affinity matrix given by the specified function and applies spectral decomposition to the corresponding graph laplacian.(D. Donoho, et al., 2003) The graph generated can be considered as a discrete approximation of the low dimensional manifold in the high dimensional space. 

The implementation of Specrtral Embedding model here is a special version, with the name Laplacian Eigenmap. The approach of Laplacian Eigenmap on dimensionality reduction is based on the intrinsic geometric structure of the manifold, which exhibits stability with respect to the embedding. And the representation will not change as long as the embedding is isometric. (M. Belkin et al., 2003)

\textbf{T-distributed Stochastic Neighbor Embedding (tSNE)}

T-SNE visualizes high-dimensional data by giving each datapoint a location in a low-dimension map.(L. van der Maaten, 2008) The technique is a variation of Stochastic Neighbor Embedding (H. Geoffery et al., 2002). It could achieve better visualizations by reducing tendency of crowding the points in the center of the map and reveals structure at many different scales. All those manifold learning algorithms can be applied here, so the plan of this project is to try a subset (if not all) of them and compare the results.

\textbf{Multidimensional Scaling (MDS)}

MDS leverages the concept of similarity, or a sense of "sameness" among things through a quantitative estimate. It is more formally referring to a set of statistical techniques that are used to reduce the complexity of a dataset.(H. Michael et al., 2013).

\textbf{Locally linear embedding (LLE)}

Locally linear embedding (LLE) seeks a lower-dimensional projection of the data which preserves distances within local neighborhoods, which can be thought of as a series of local Principal Component Analyses which are globally compared to find the best non-linear embedding (T, Joshua et al., 2000).

\textbf{ISOMAP}

ISOMAP is one representative of isometric mapping methods, and extends metric multidimensional scaling (MDS) by incorporating the geodesic distances imposed by a weighted graph.(wikipedia) As we discussed in the class, the key idea of ISOMAP is to produce low dimensional representation which preserves ``walking distance" over the data cloud (manifold). 

\subsection{Data Transformation}

The Data Transformation steps are one of the most important and heavy-lifting stage in the neural network training and processing stage. As mentioned before, the standard input data format for most neural network models are tensors. Within each datapoint (each image), the shape need to be $224 \times 224$, which not all artworks are presented with this shape in their numpy array forms. 

To perform standardization of input data, each image, regardless of sizes, need to be reformatted to $224 \times 224$ pixels. A square pad function is applied to each image to give each image a padding if the image is not square, and then resize the image to the required pixel format. During this stage, it appears that some image sizes are too large, hence some down sampling and center cropping were applied to make sure the image data satisfies the input requirements meanwhile preserve the original qualify of the picture as much as possible.

Additionally, some artworks are just sketches and not satisfying the dimension requirements as input tensors, hence those artworks will be dropped as bad data.

Finally, the full image data was transformed to a tensor with shape of (3198,3,224,224) for future processing. This data tensor, along with the index locations of bad data, are saved as pickle files for future use. 

This concludes the initial data preprocessing and transformation stage and the data is now ready to go to be passed into neural networks.

\subsection{Initial Implementation} 

\textbf{ResNet18}

As a starting point, RstNet18, the smallest and most lightweight implementation in ResNet family with 11689512 parameters and 18 layers, is explored just to visualize the data and obtain a general sense of what ResNet would be capable of. Note that the art categories are not merged yet in this implementation. 

\begin{figure}[!ht]
    \centering
    \subfigure[ResNet18-tSNE]{\includegraphics[width=0.22\textwidth]{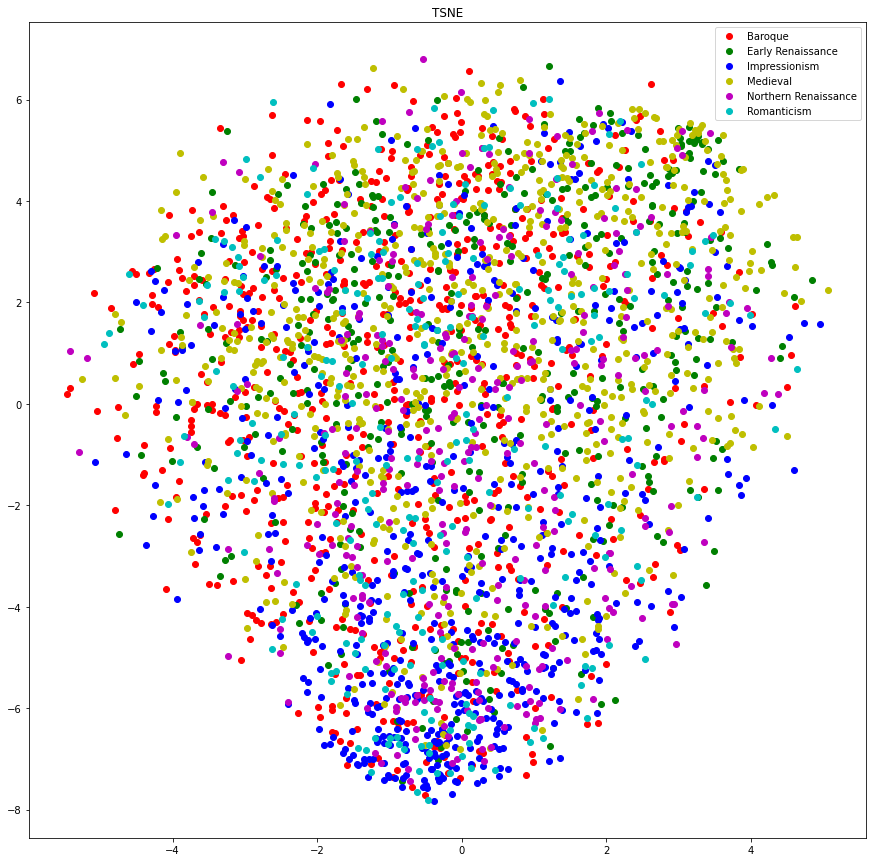}} 
    \subfigure[ResNet18-MDS]{\includegraphics[width=0.22\textwidth]{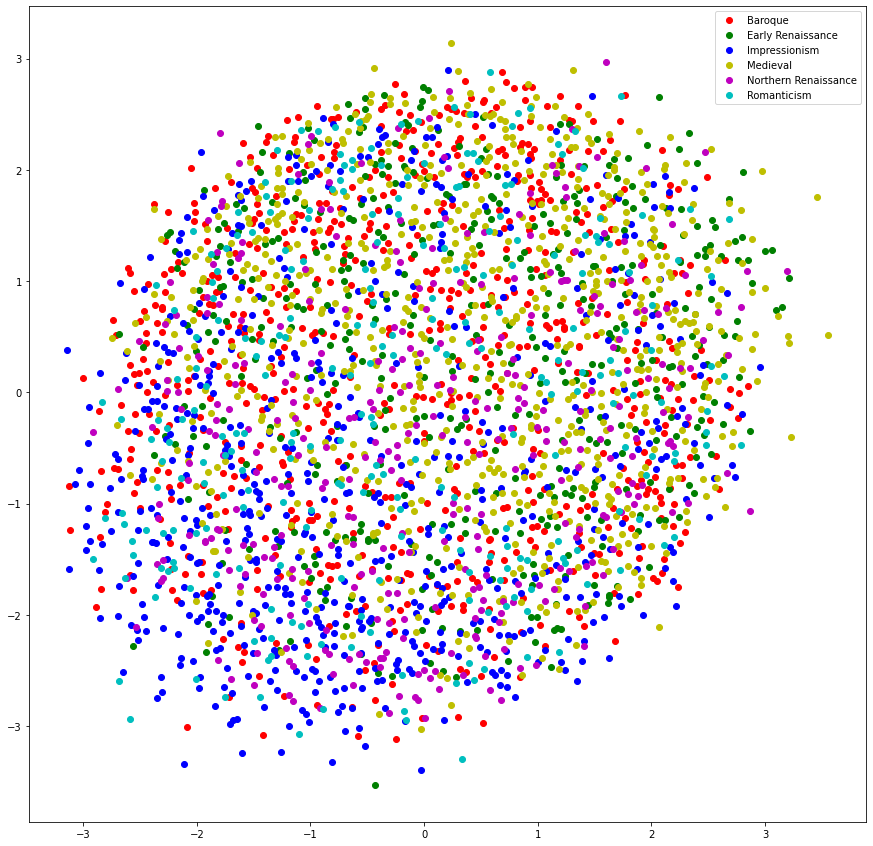}} 
    \subfigure[ResNet18-ISOMAP]{\includegraphics[width=0.22\textwidth]{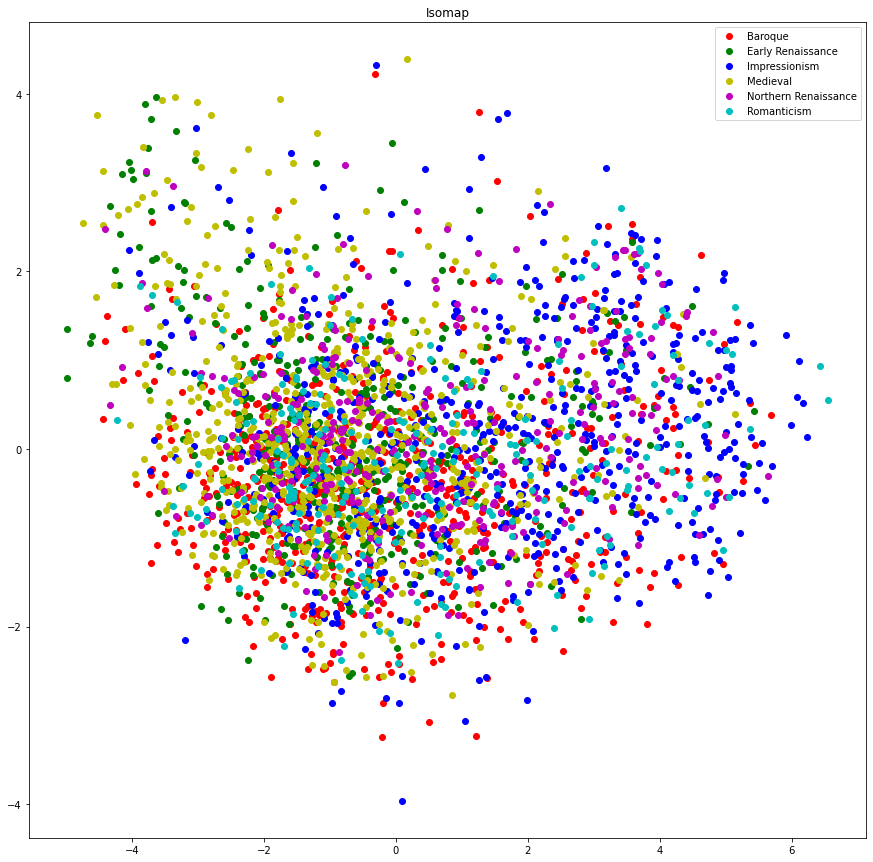}}
    \subfigure[ResNet18-LLE]{\includegraphics[width=0.22\textwidth]{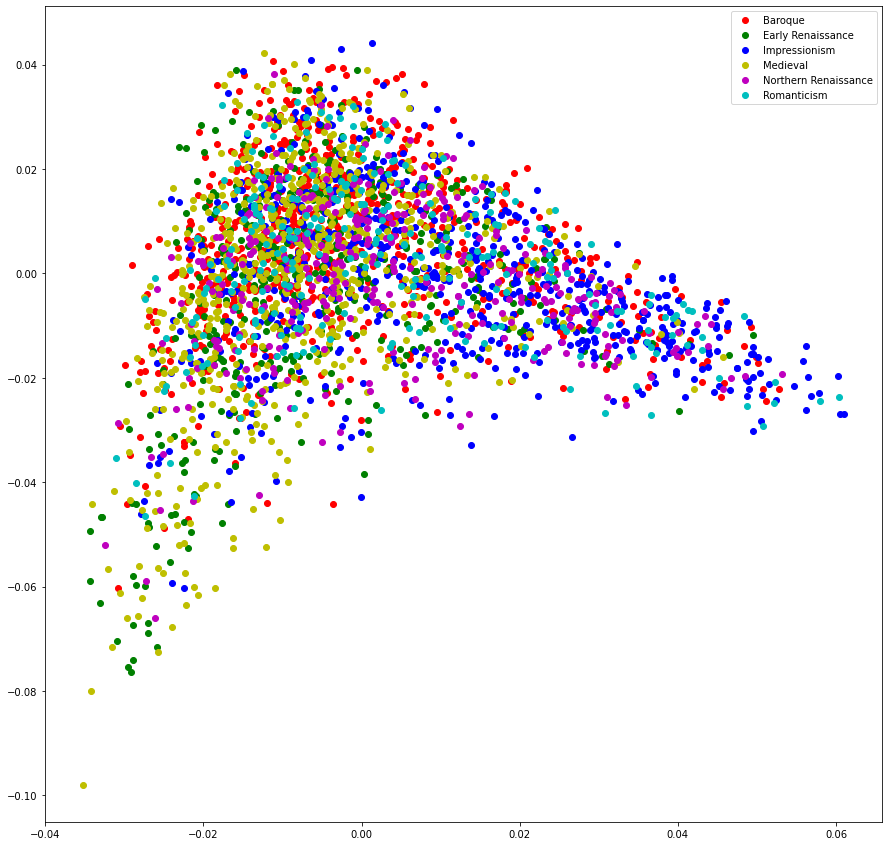}} 

    \caption{ResNet18 Visualized}
    \label{fig:resnet18}
\end{figure}

As shown from Figure \ref{fig:resnet18}, Resnet18 is not doing a very well job identifying and classifying different art categories - The entire visualization done by tSNE, MDS, ISOMAP tend to look like a big ball and all the colors representing different art categories are mixed together despite there are very subtle grouping of impressionism. 

This exploration is telling us that in the eyes of machines applied here, there is little or no differentiation between art categories, not to mention the ``flow" of art history.

After this exploration, the Early Renaissance and Northern Renaissance categories are merged into one category named Renaissance given the similarity of that two groups are relatively close. 

\clearpage

\textbf{Resnet50} 

ResNet50 is a deeper ResNet variant than ResNet18 with number of parameters = 25557032 and 50 layers. Then tSNE, ISOMAP, Laplacian Eigenmap and LLE are applied to the feature embedding to visualize the data. 

\begin{figure}[!ht]
    \centering
    \subfigure[ResNet50-ISOMAP]{\includegraphics[width=0.22\textwidth]{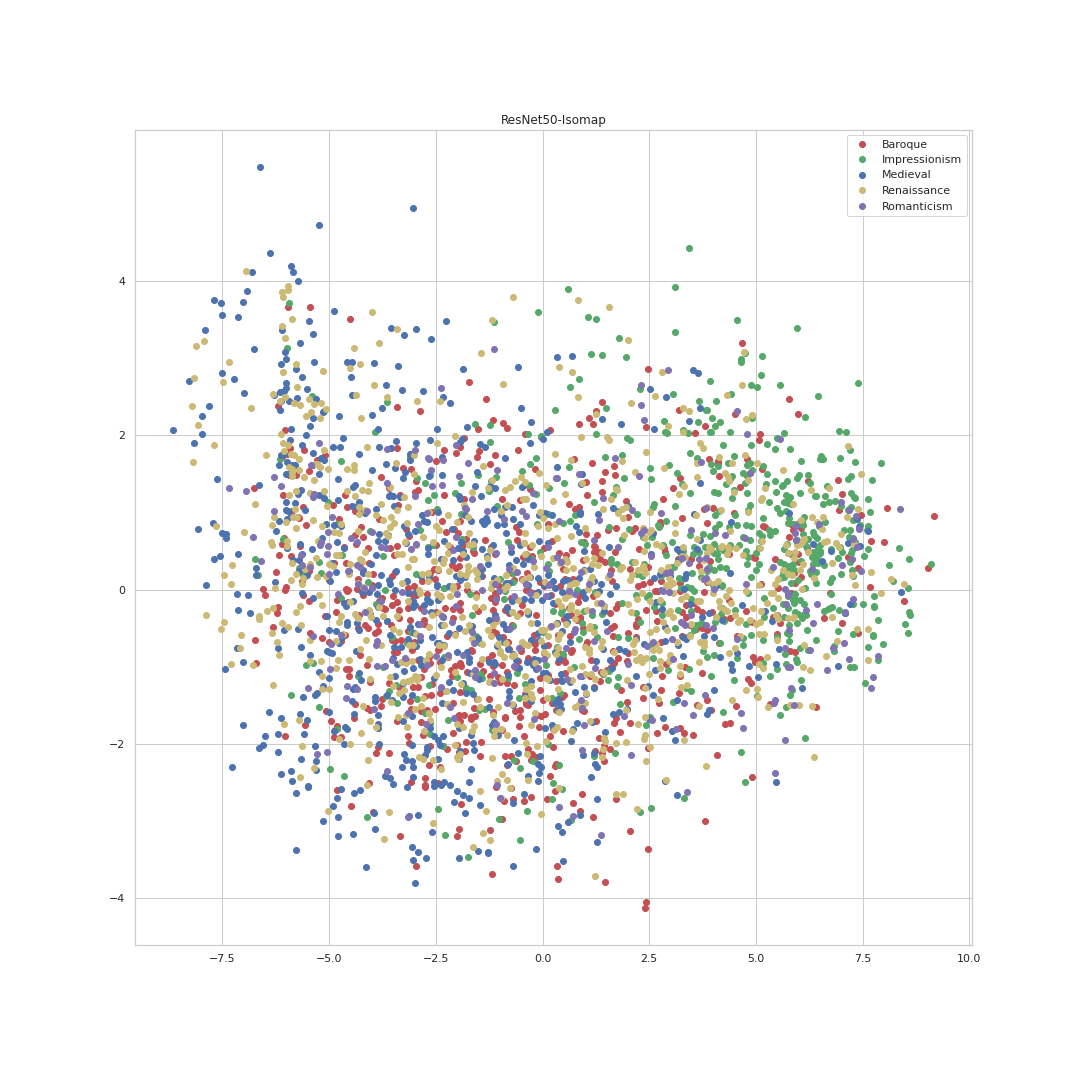}} 
    \subfigure[ResNet50-tSNE]{\includegraphics[width=0.22\textwidth]{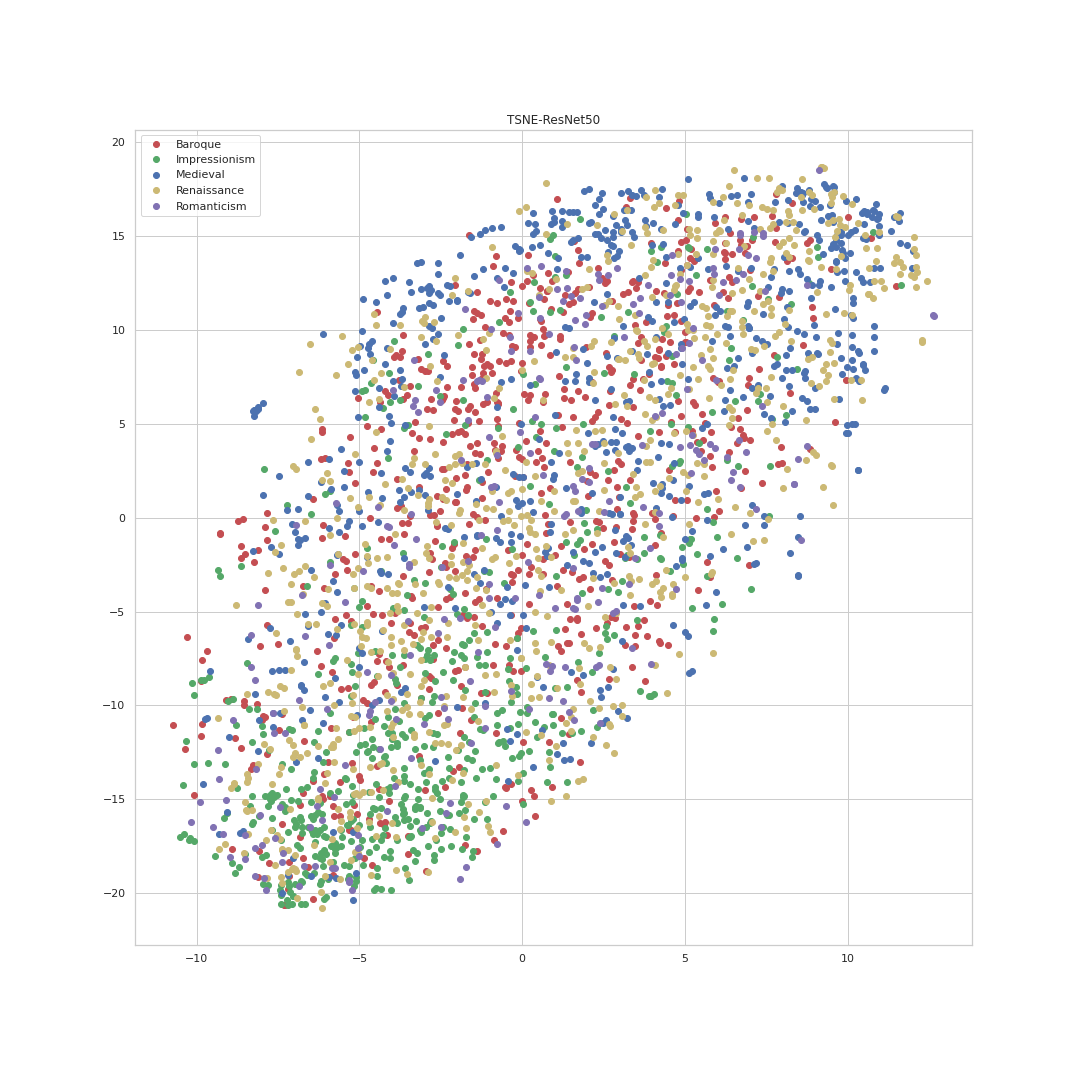}} 
    \subfigure[ResNet50-Laplacian Eigenmap]{\includegraphics[width=0.22\textwidth]{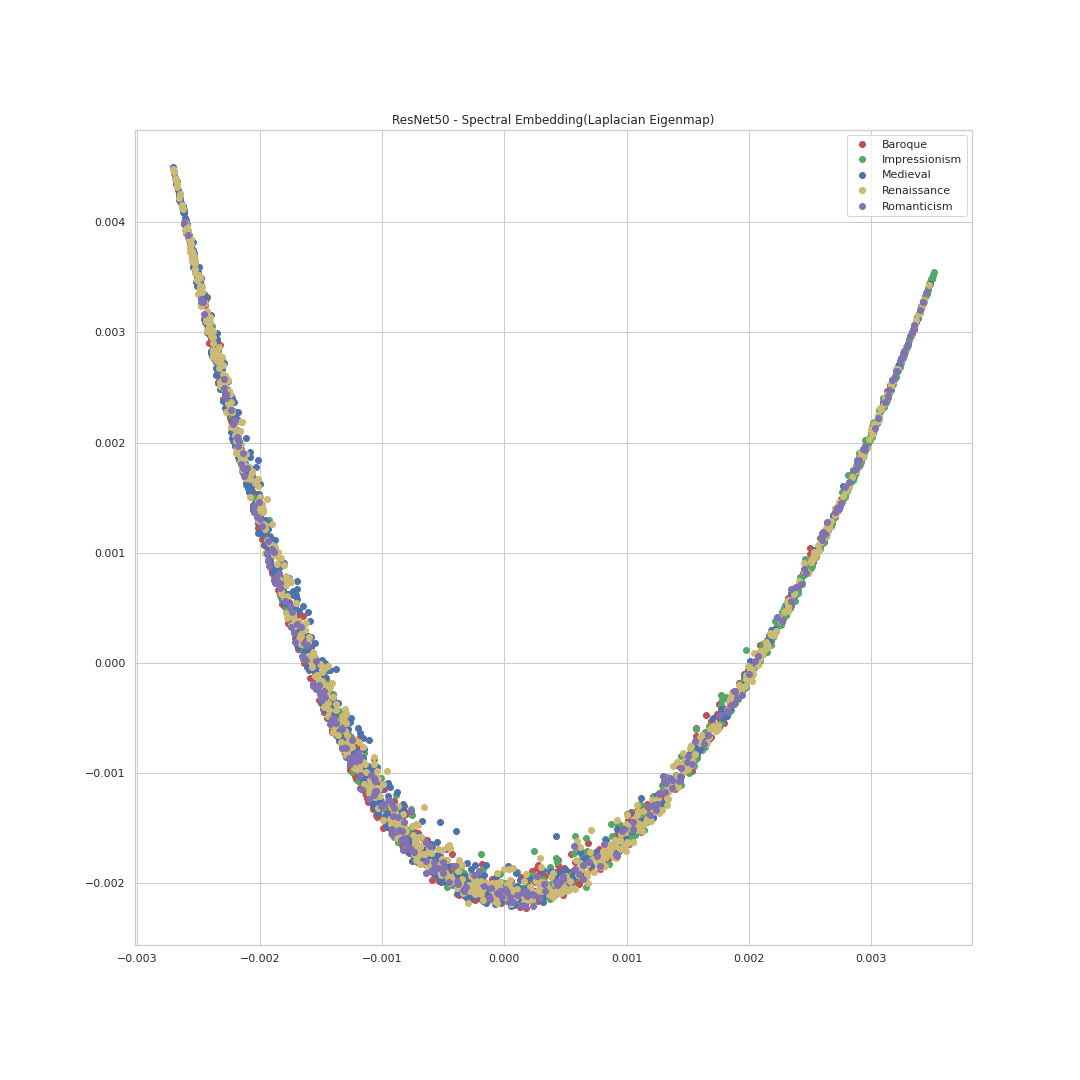}}
    \subfigure[ResNet50-LLE]{\includegraphics[width=0.22\textwidth]{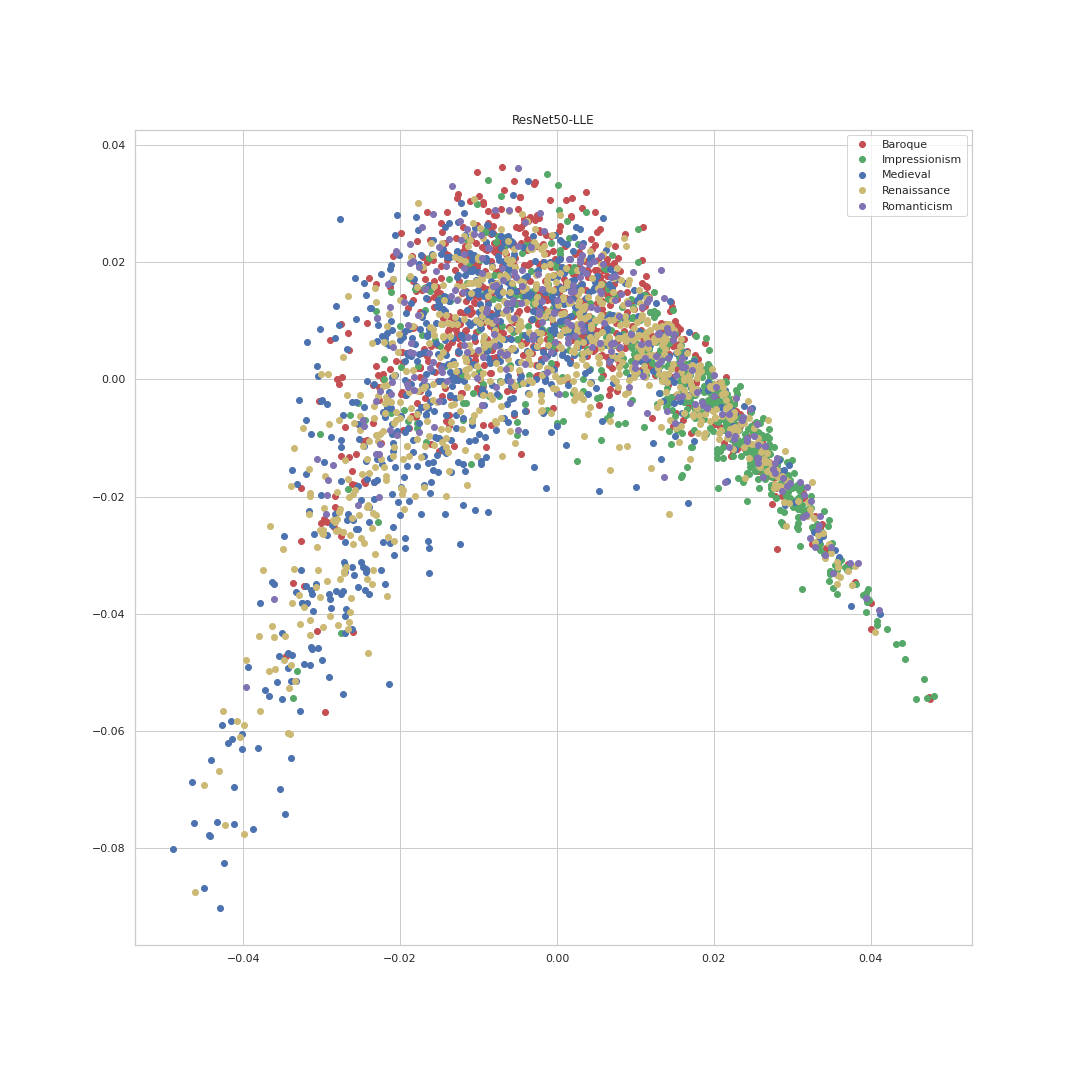}} 

    \caption{ResNet18 Visualized}
    \label{fig:resnet50}
\end{figure}

As shown in \ref{fig:resnet50} The result does improve much from ResNet18, which is not intuitive since there no proven better performance of a aggressively deep model over shallow model. During the RestNet training and exploration process, the testing result of a 1202-layer network could is worse than a 110-layer network. (Kaiming H et al., 2015). Given the cpu and local environment constraints, it's wiser to switch to other nerasl network models and see better feature embedding can be achieved.

\textbf{AlexNet}

AlexNet was introduced earlier than RestNet and was the award-winner of the 2012 ImageNet contest. Despite all the attractiveness of CNNs and their relative efficiency of local architecture, they are still relatively expensive to apply in large scale, high-resolution images (A. Krizhevsky et al., 2010), especially in my current local environment. This project applied pre-trained weights of AlexNet to construct the feature embedding and then visualized through ISOMAP, Eigenmap, LLE, tSNE. 

\begin{figure}[!ht]
    \centering
    \subfigure[AlexNet-ISOMAP]{\includegraphics[width=0.22\textwidth]{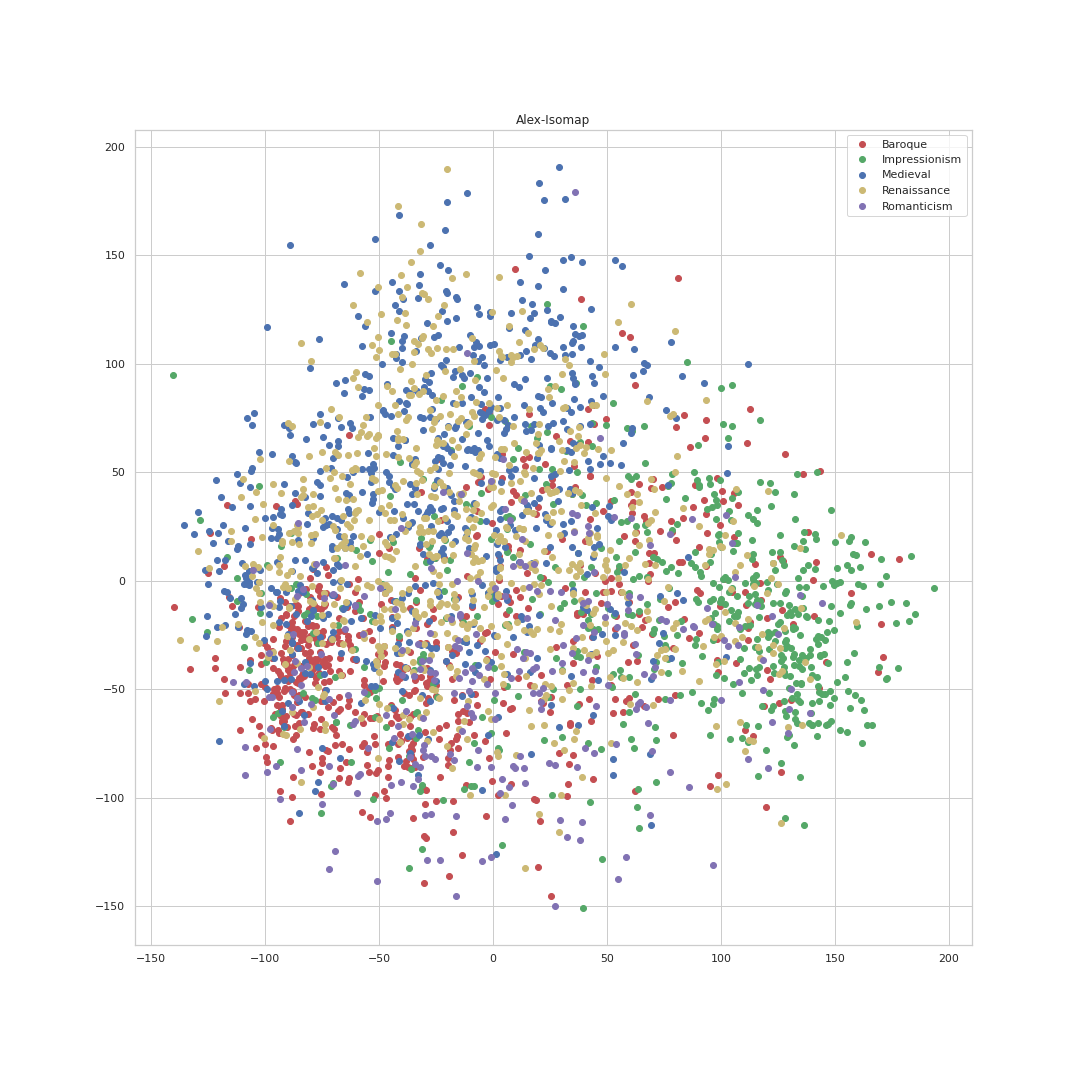}} 
    \subfigure[AlexNet-Laplacian Eigenmap]{\includegraphics[width=0.22\textwidth]{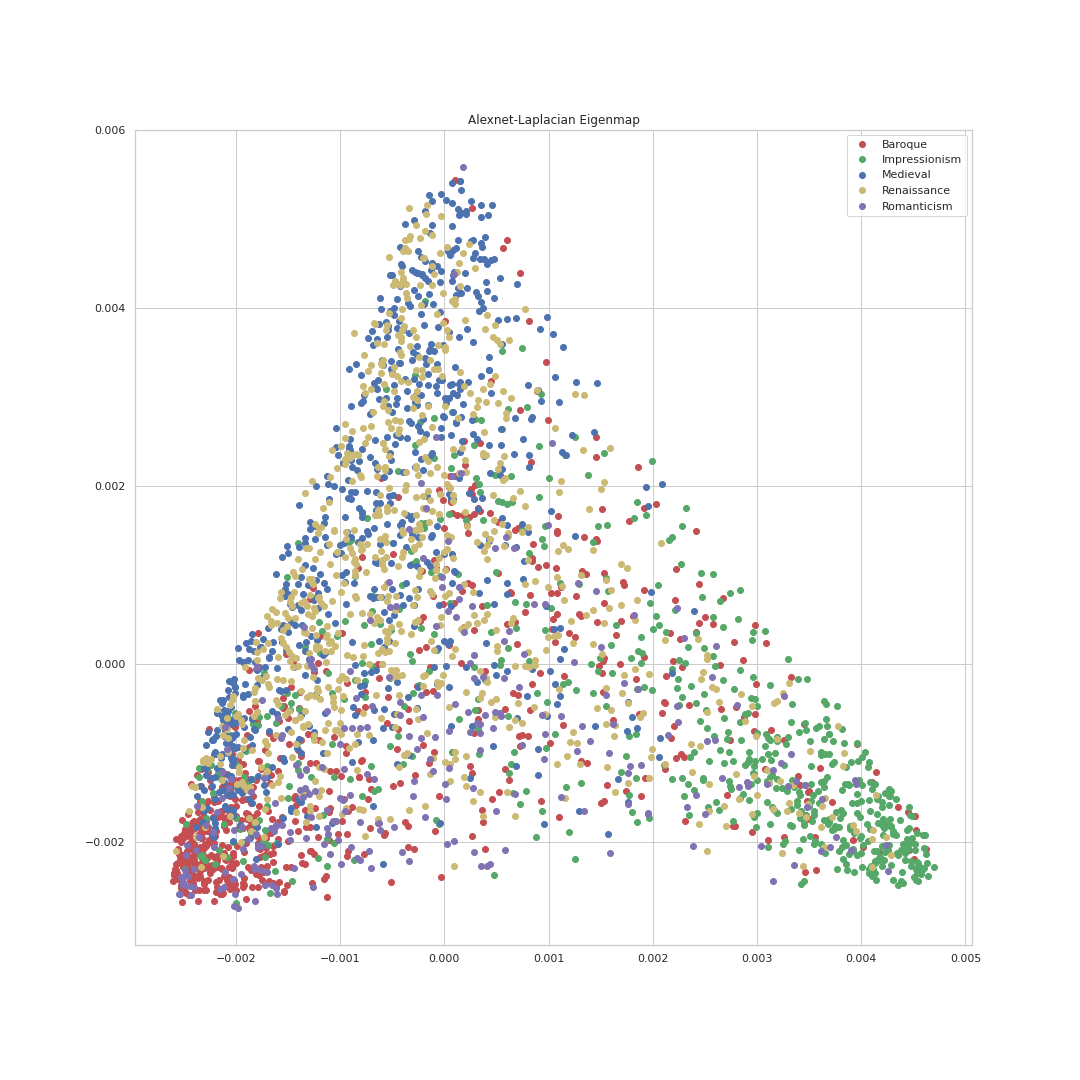}} 
    \subfigure[AlexNet-LLE]{\includegraphics[width=0.22\textwidth]{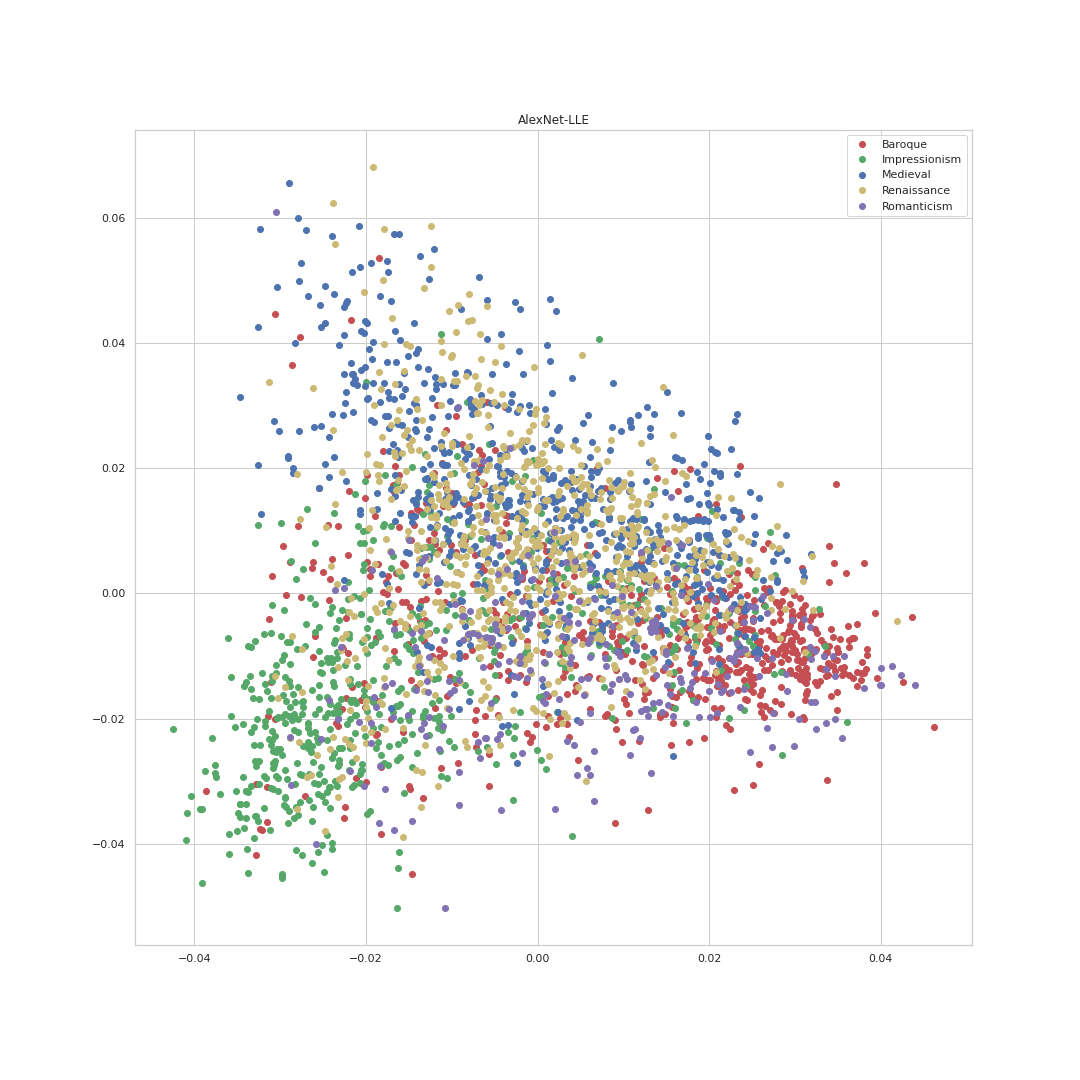}}
    \subfigure[AlexNet-tSNE]{\includegraphics[width=0.22\textwidth]{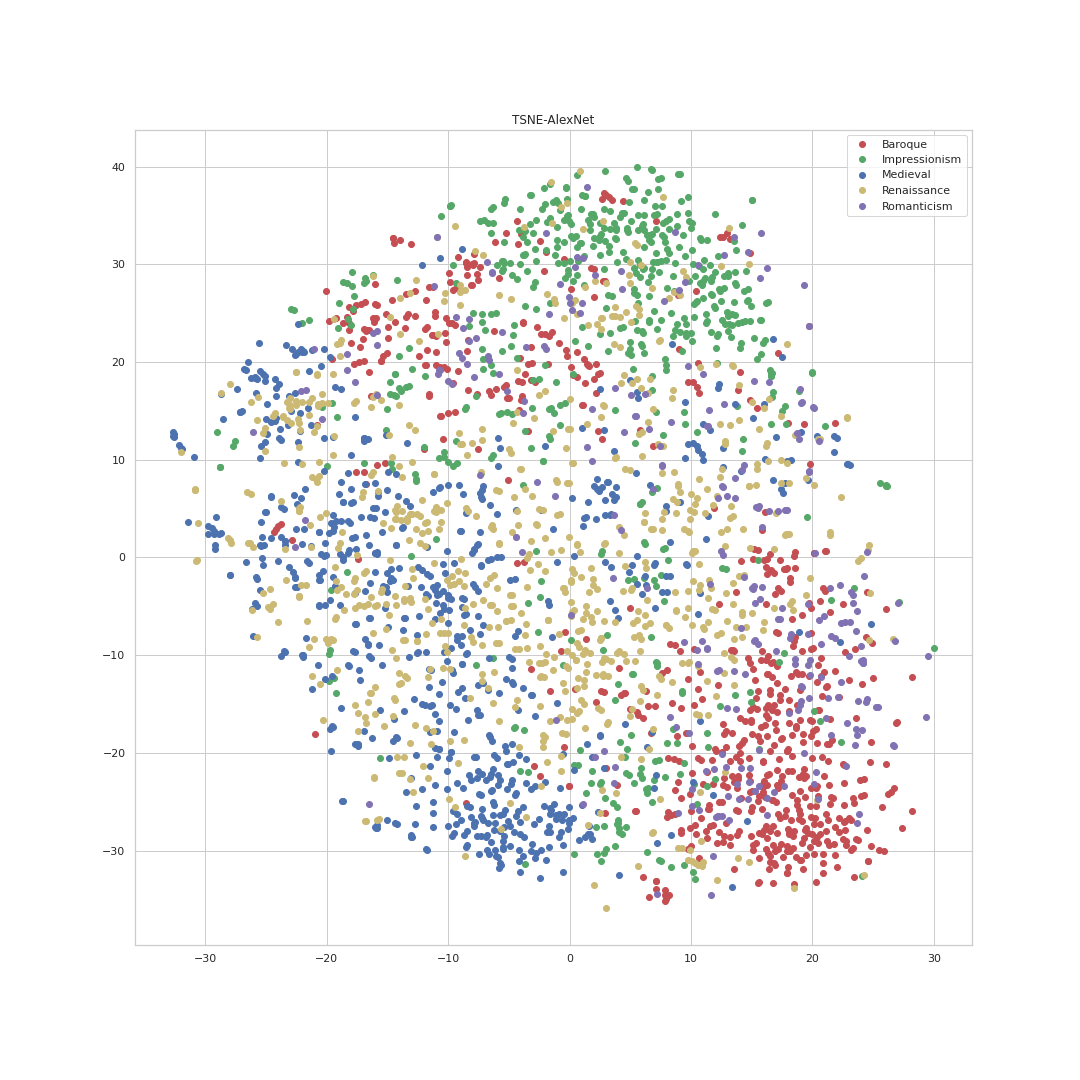}} 

    \caption{AlexNet Visualized}
    \label{fig:alexnet}
\end{figure}

From Figrues \ref{fig:alexnet}, there is clearly a better performance achieved by AlexNet comparing to ResNet family that the image datapoints are starting to show clusters. The blue points represent Medieval, red points represent Baroque and green points represent impressionism, which are the considered very stylish and representing a period of their own era. For tSNE representation, there are smaller clusters that could be observed far away from their main cluster. It could be some images that have belong to one category but looks like another in terms of styles, which will be further explored in this project.

\textbf{Vision Transformers}

Vision Transformer is the latest introduced neural network within the selected model candidates in this project. Two different variants: ViT16 and ViT32 are tried to perform the feature embedding and similar, various manifold learning algorithms are applied to the embedded feature and reduce the dimension to 2-d so they can be visualized. 

In Laplacian Eigenmap and LLE, there is a rich group of blue dots on the lower side representing the Medieval period. Then the gold dots representing Renaissance was spanning from Medieval period to Baroque period described by red dots indicating it's a revolutionized movement in art styles. The scattered purple dots representing Romanticism in between baroque and impressionism can barely be observed because of its mixed art nature and relative low volume of data. Finally the the green dots representing impressionism started from roughly where Romanticism lays and gradually become dominant and with more distinct style over the time. 


Additionally, from the feature embedding of ViT16 to ViT30, the visual representation of ISOMAP and LLE looks more different than the one done by tSNE and Laplacian Eigenmap under similar parameters. ISOMAP seems achieve better clusters under similar parameters and the shape of LLE turned from a flow to a triangle. It is hard to determine through eyeballing which one outperformed and since the next sections of this project will focus more on Laplacian Eigenmap and PHATE algorithm (which will be introduced later), Vit16 model is chosen to move forward for a more lightweight implementation. 

We will look deeper into more details in the next sections.

\begin{figure}[!ht]
    \centering
    \subfigure[ViT16-ISOMAP]{\includegraphics[width=0.22\textwidth]{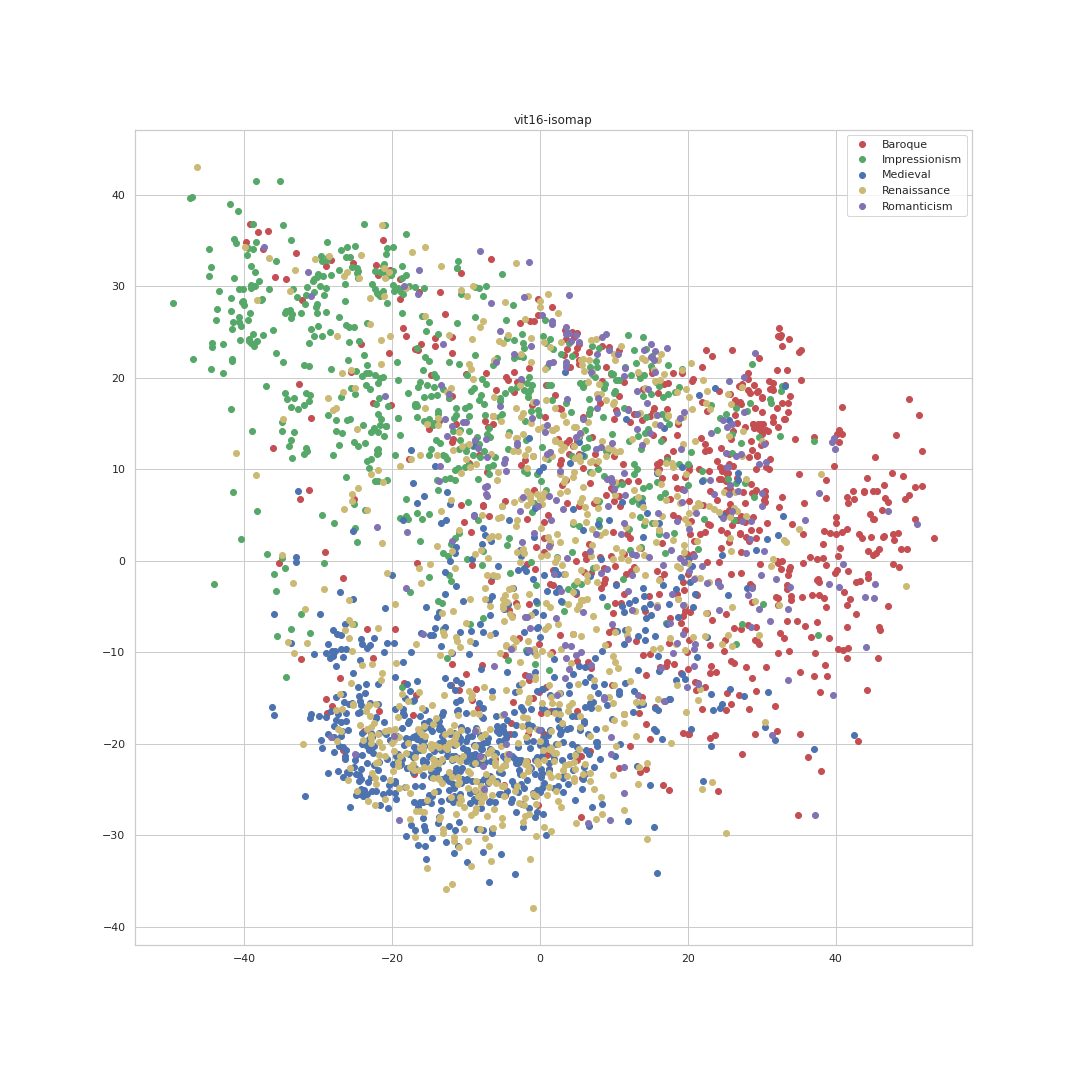}} 
    \subfigure[ViT16-LLE]{\includegraphics[width=0.22\textwidth]{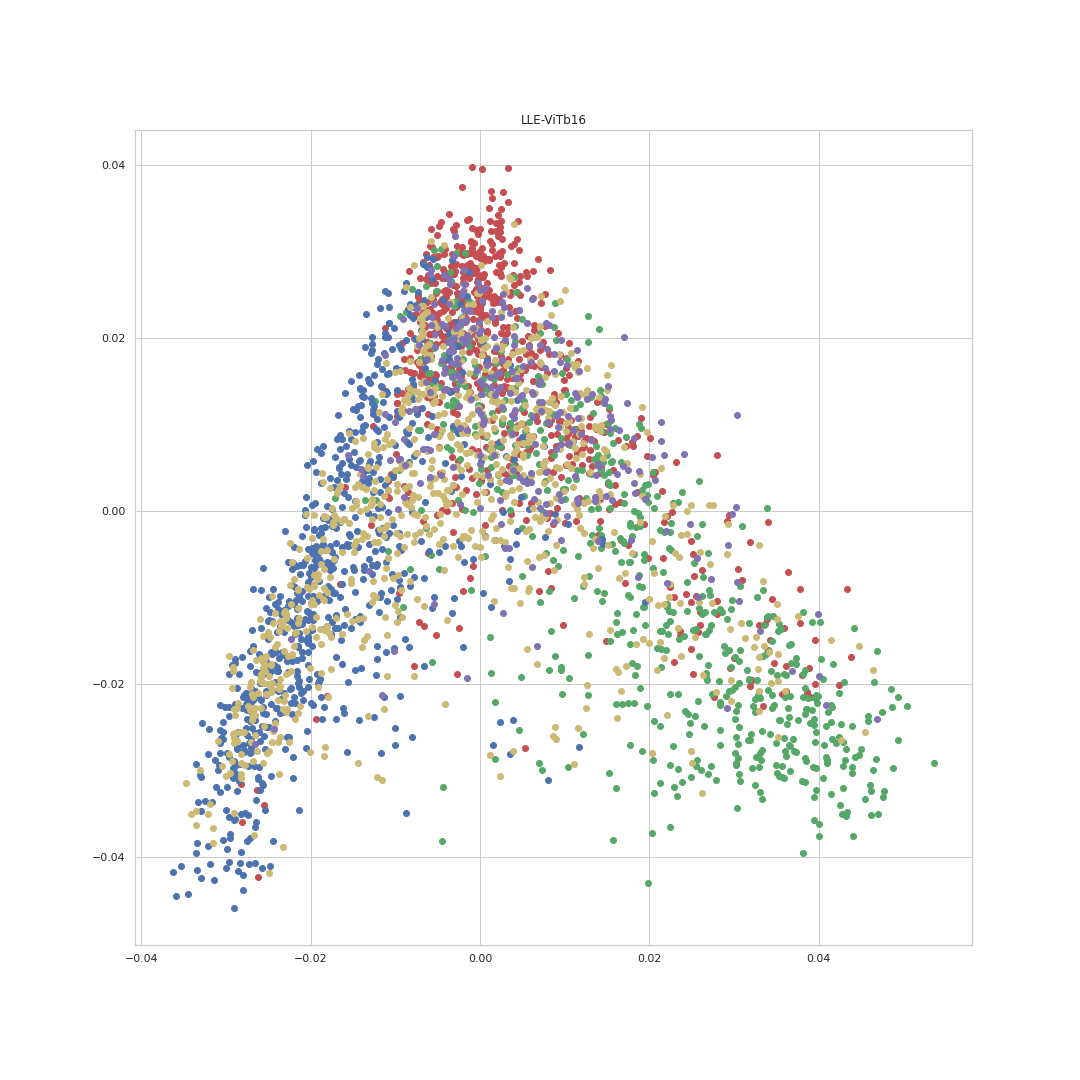}}
    \subfigure[ViT16-Laplacian Eigenmap]{\includegraphics[width=0.22\textwidth]{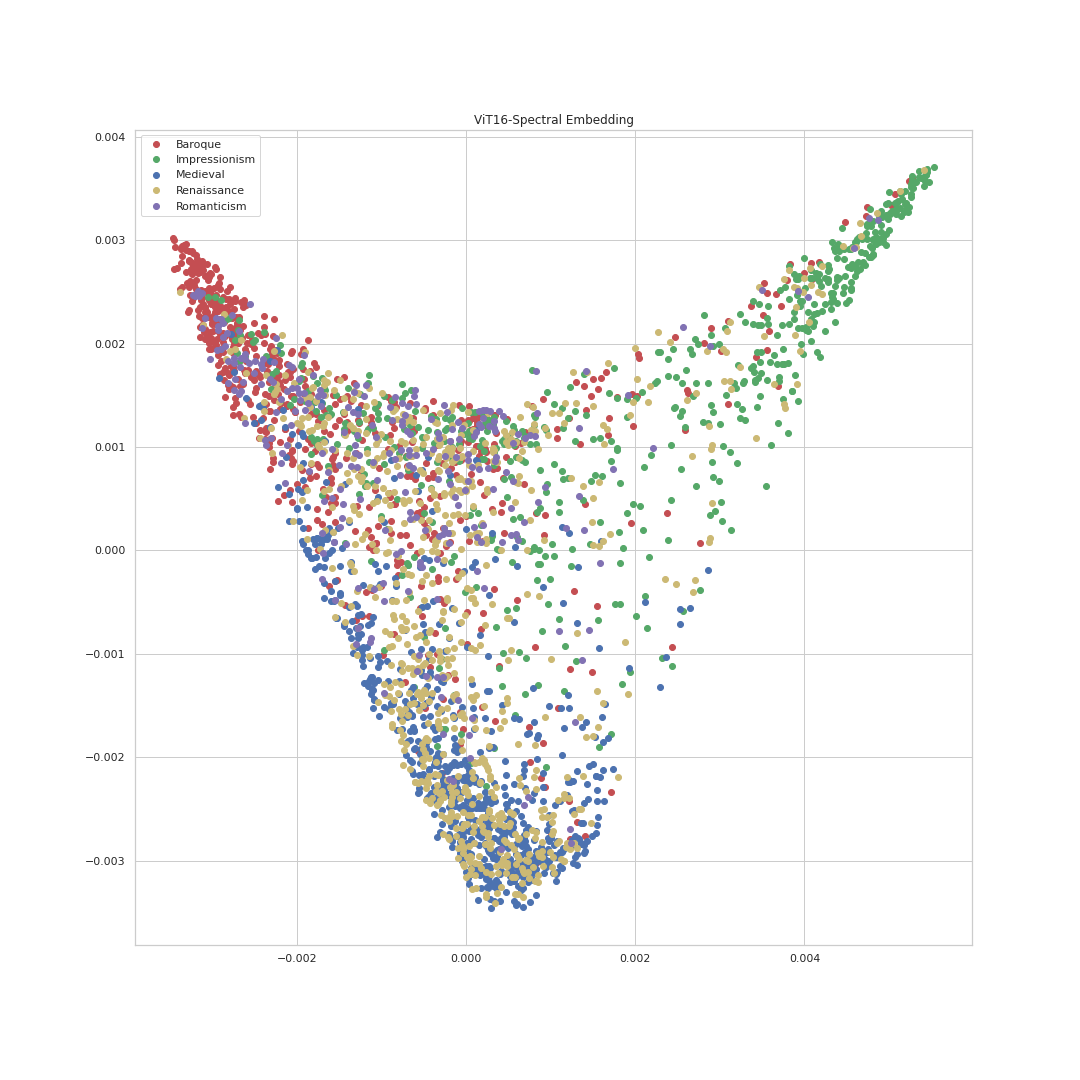}} 
    \subfigure[ViT16-tSNE]{\includegraphics[width=0.22\textwidth]{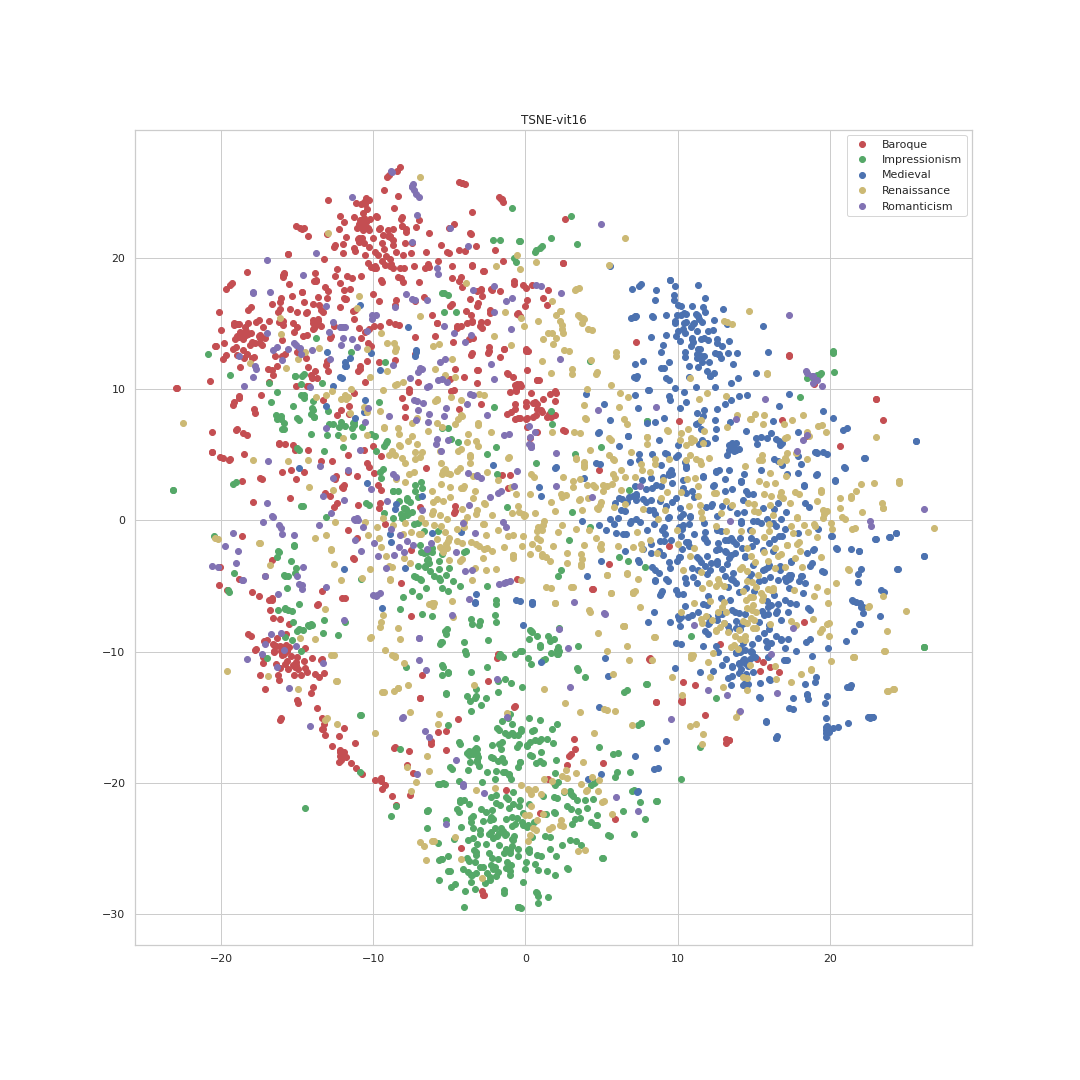}} 

    \caption{Vision Transformer 16 Visualized}
    \label{fig:vit16}
\end{figure}

\begin{figure}[!ht]
    \centering
    \subfigure[ViT32-ISOMAP]{\includegraphics[width=0.22\textwidth]{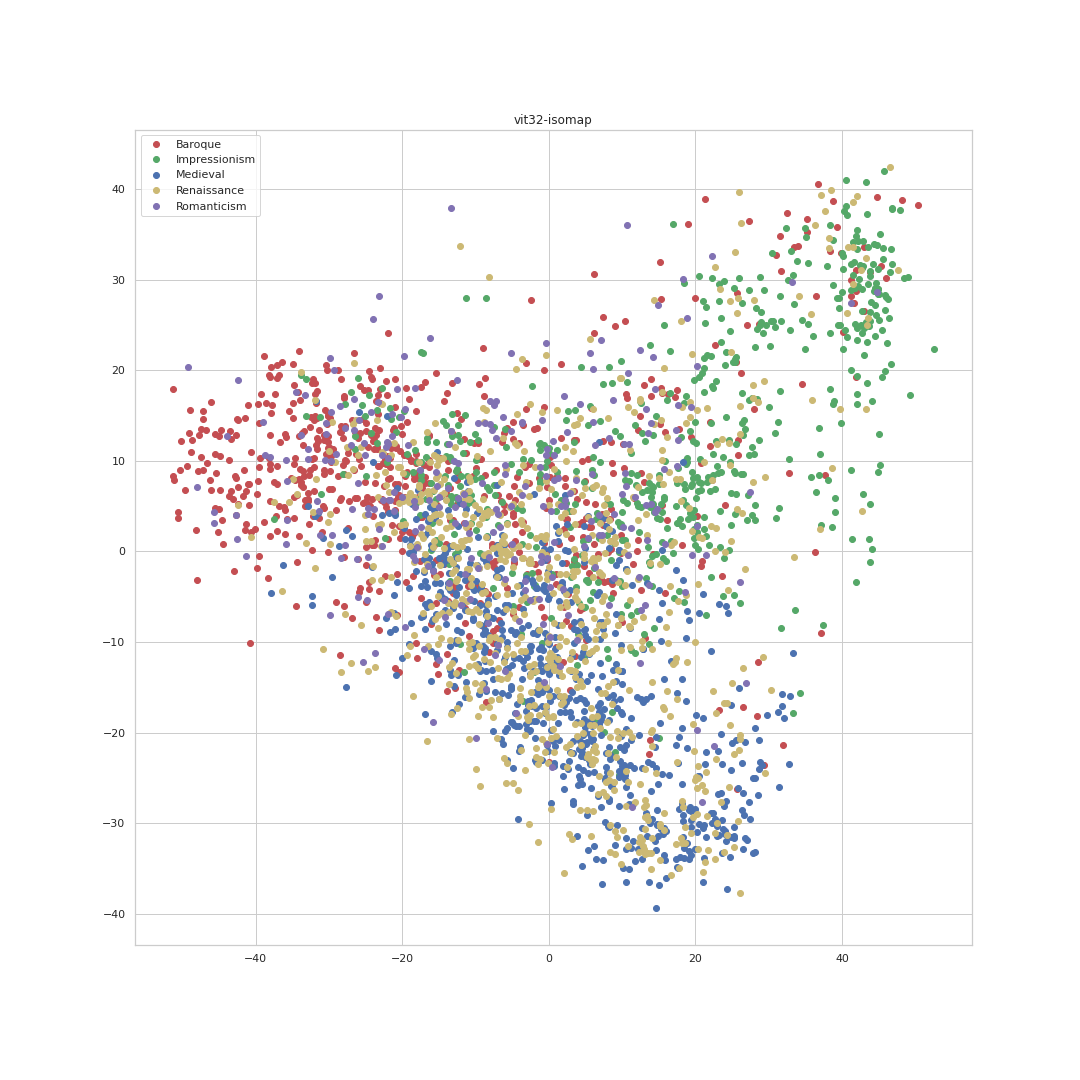}} 
    \subfigure[ViT32-LLE]{\includegraphics[width=0.22\textwidth]{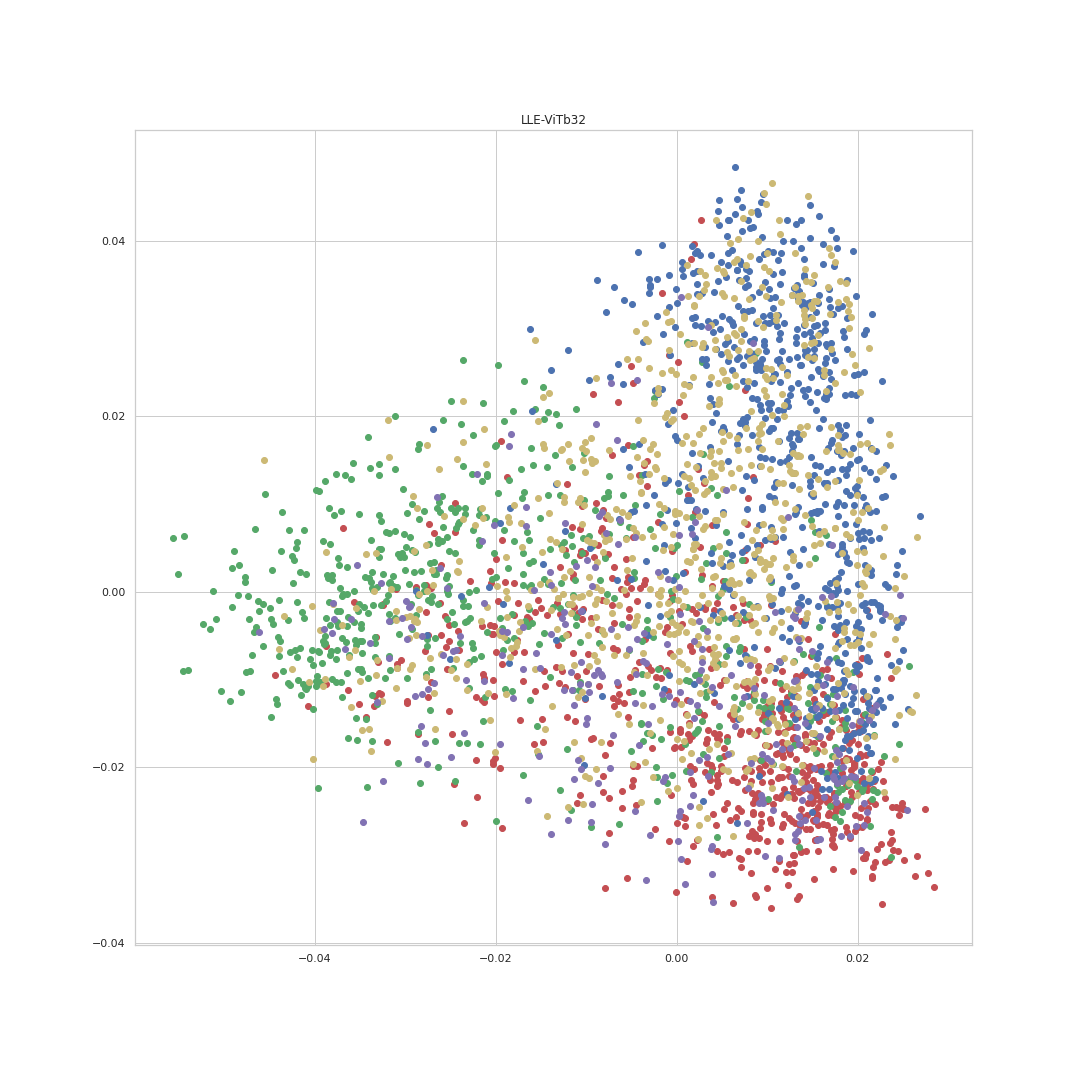}} 
    \subfigure[ViT32-Laplacian Eigenmap ]{\includegraphics[width=0.22\textwidth]{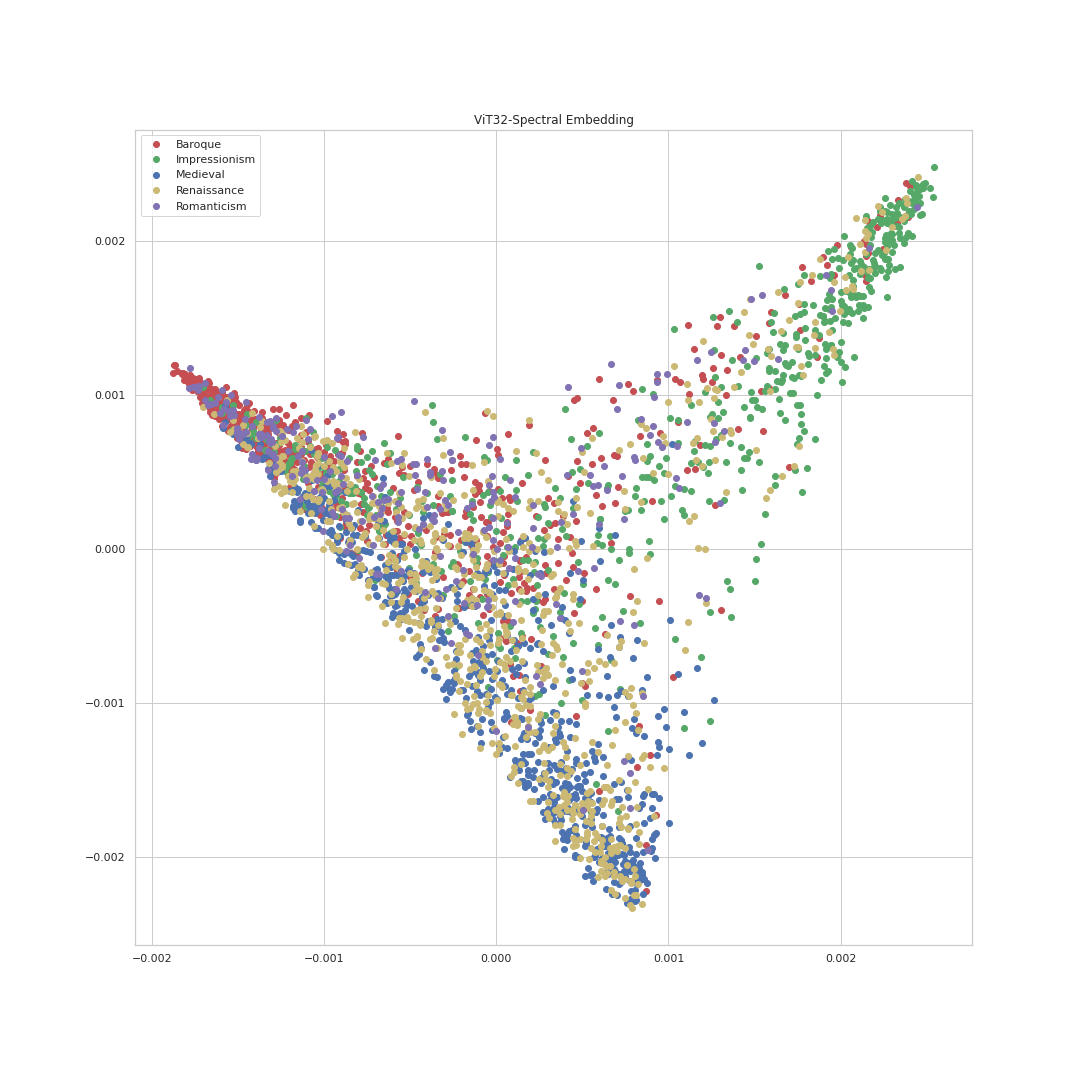}}
    \subfigure[ViT32-tSNE]{\includegraphics[width=0.22\textwidth]{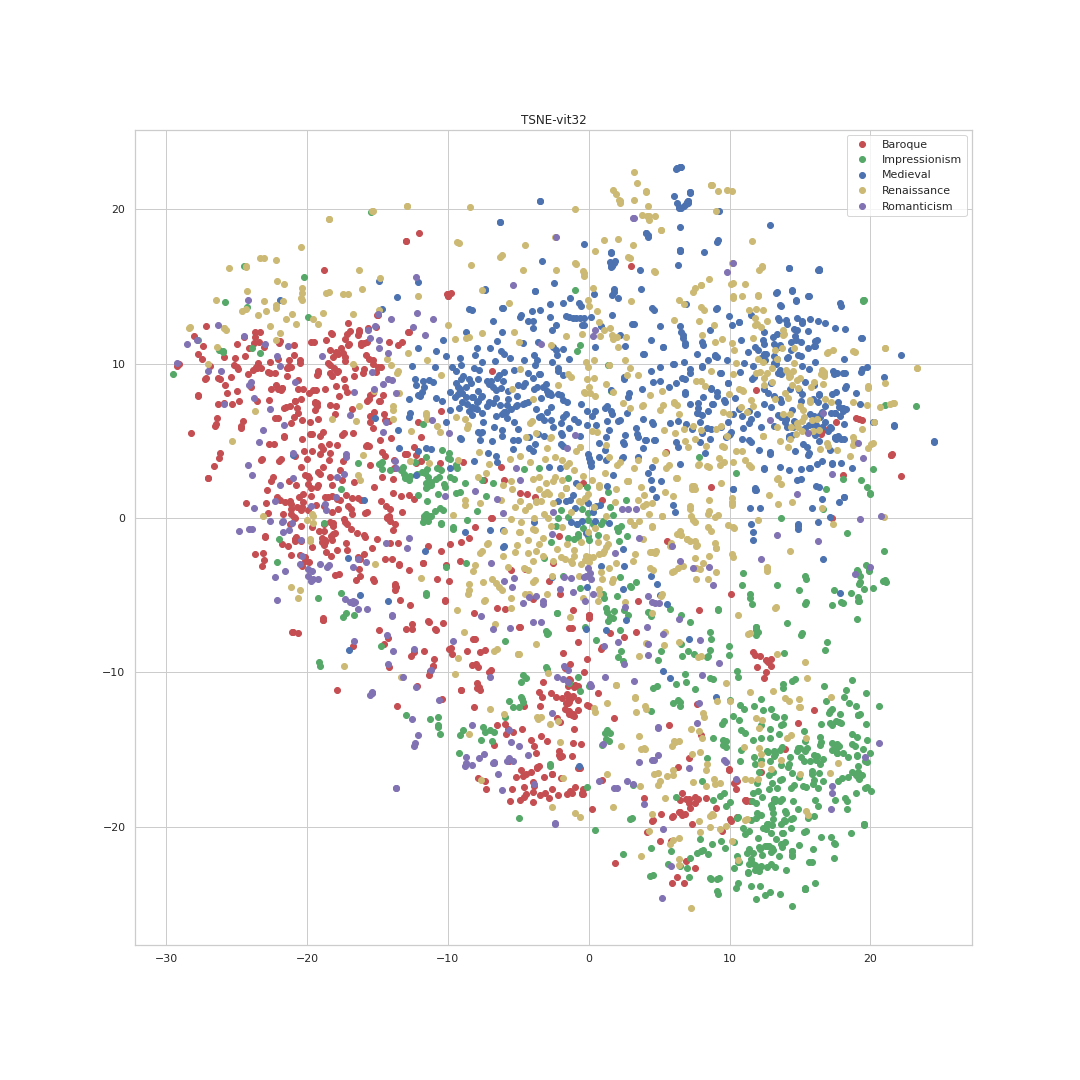}} 

    \caption{Vision Transformer 32 Visualized}
    \label{fig:vit32}
\end{figure}

\clearpage

\subsection{Evaluation and Interpretation}

The following table summarizes the rating and comparison of each neural network and manifold algorithm pairs. 
\begin{center}
\begin{tabular}{|c|c|c|c|c|}
\hline
\textbf{Neural Network} & \textbf{tSNE} & \textbf{Laplacian Eigenmap} & \textbf{ISOMAP} & \textbf{LLE} \\\hline
    ResNet16 & $\star$ & $\star$ & $\star$  & $\star\star$ \\\hline
    ResNet50 & $\star$ & $\star$ & $\star$ & $\star\star$ \\ \hline
    AlexNet & $\star\star$ & $\star\star$ & $\star\star$ & $\star\star$ \\ \hline
    ViT16 & $\star\star\star$ & $\star\star \star \star$ & $\star \star \star$ & $\star \star \star \star$ \\ \hline
    ViT32 & $\star \star$ & $\star  \star \star$ & $\star \star \star$ & $\star \star \star$ \\ \hline
\end{tabular}
\end{center}

As shown in Figure \ref{fig:vit16} and \ref{fig:vit32}, the Vision Transformer models achieved a better performance than AlexNet and ResNet in terms of image recognition and feature embedding. However, slightly different from expected,  ViT32 did not achieve a significantly better performance than ViT16, further indicated that neural network is not guaranteed to be better when the number of layers and the complexity increases. 

Each manifold learning algorithm has different styles and interpretation of the data cloud - tSNE tends to look for clusters and branches; ISOMAP and Laplacian Eigenmap tend to find global connectivity, which Laplacien Eigenmap did a better job preserving the geometry structure on a global perspective; LLE focused more on local conductivity rather than global and it is also doing a great job representing the flow of the history.

Among manifold learning algorithms described above, Laplacian Eigenmap and LLE outperformed others in this specific use case by observing the visualization. tSNE is good at finding clusters, yet we are looking for the evolving process of art history to explore how the local \& global connectivity of the entire dataset form a sense of ``flow", which is better captured by LLE and Laplacian Eigenmap. 

Since ViT16 is a winner from the overall ratings above, the rest of this project will focus on ViT16 for analysis, future training tasks and interpretation.

 \textbf{tSNE - Clustering and Branching} 
 
\begin{figure}[!ht]
    \centering

    \subfigure[ViT16-tSNE-Artist]{\includegraphics[width=0.4\textwidth]{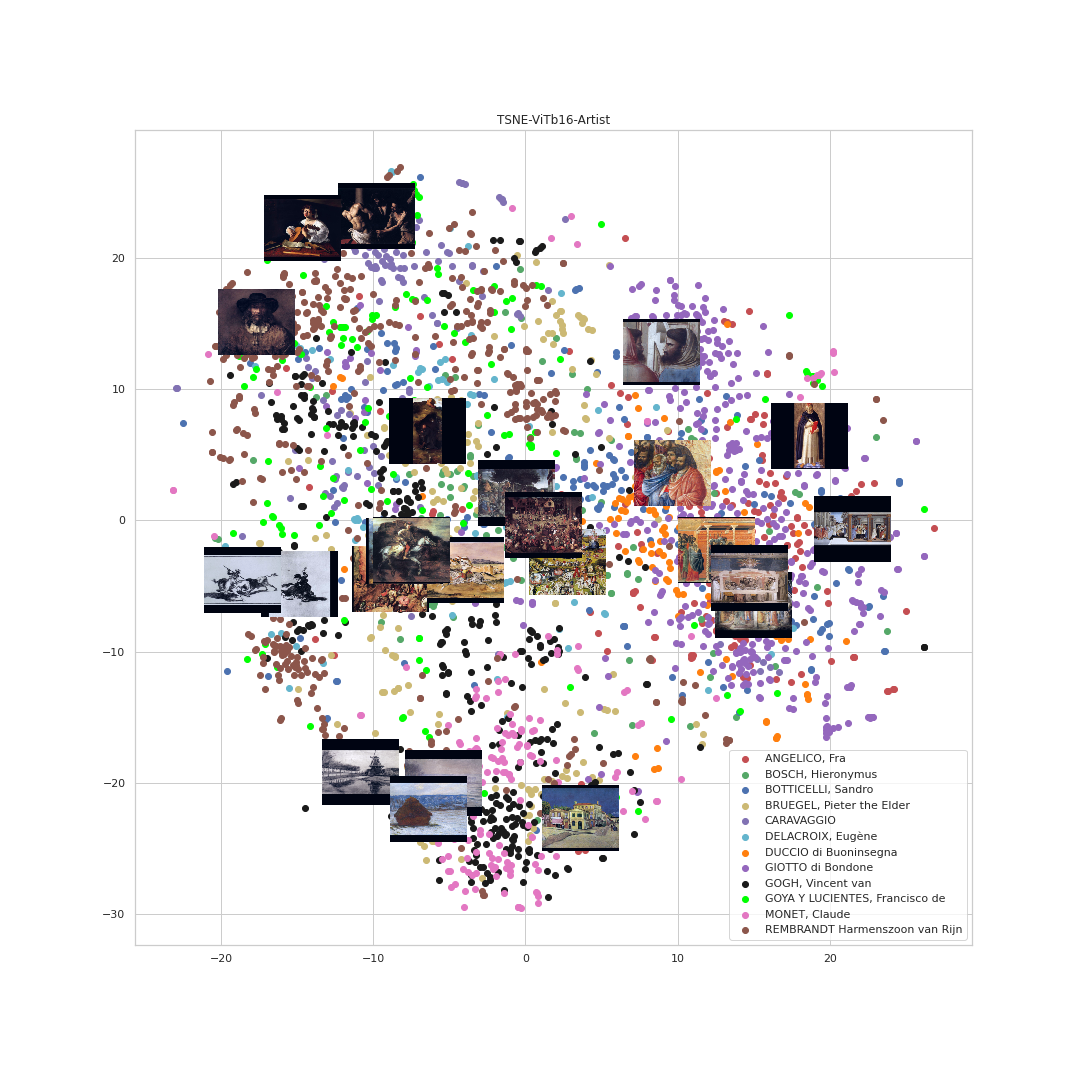}} 
    \subfigure[ViT16-tSNE-Period]{\includegraphics[width=0.4\textwidth]{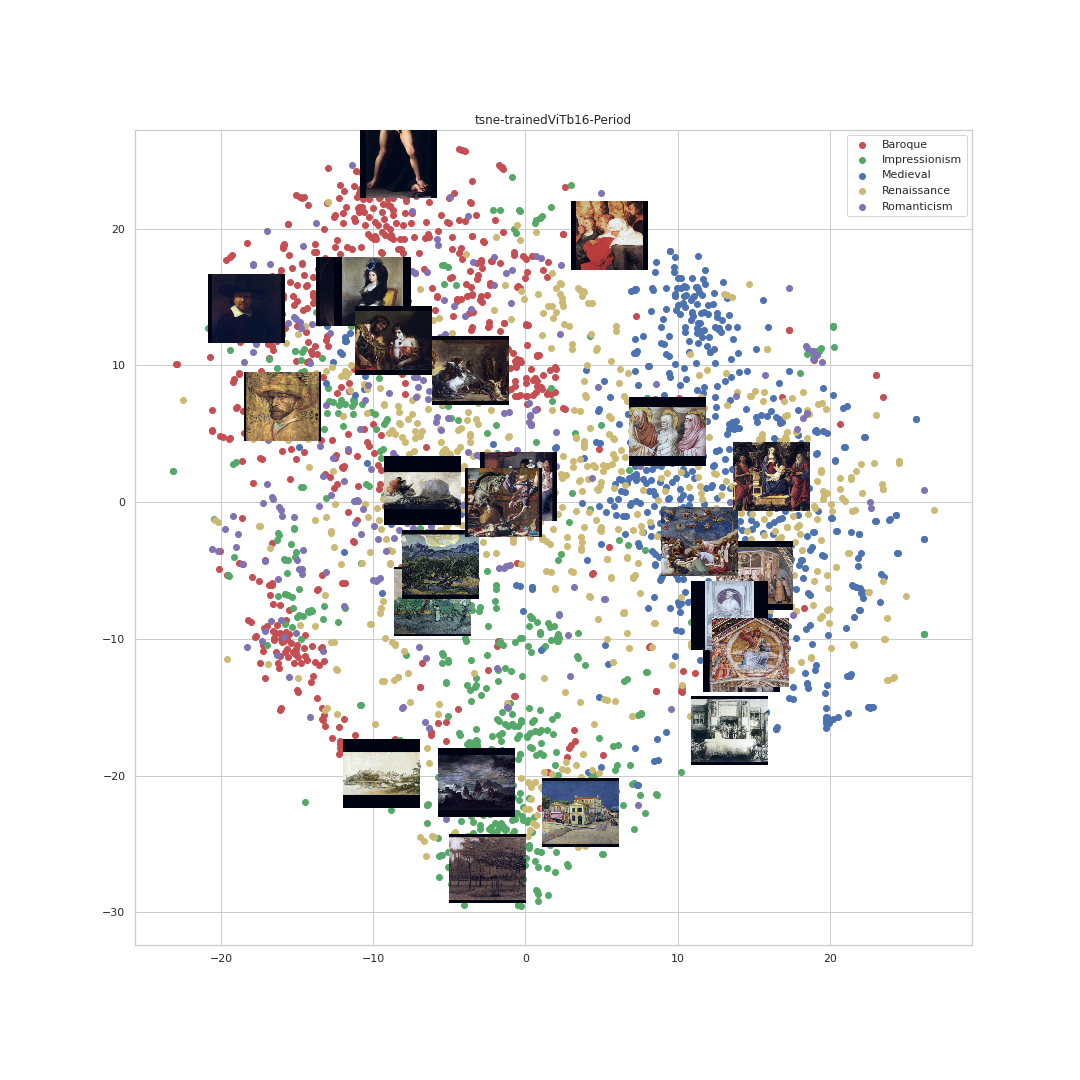}} 
    \caption{ViT16-tSNE}
    \label{fig:vit16-2}
\end{figure}

 Figure \ref{fig:vit16-2} shows a collection of random selected original images of the artwork corresponds to their location in the feature embedding. The top left side is dominated by baroque style, which flourishes in $17^{th}$ century Europe and establishes strong contrast, deep color and rich detail to achieve a sense of awe. Baroque was the principal European style in the visual arts of the $17^{th}$ century, covers various national styles from the complex and dramatic Italian art of the $17^{th}$  century to the restrained genre scenes, still-lifes and portraits characteristic of the Dutch Baroque.  (oxfordartonline) 
 
 In the annotation dataset, Caravaggio and Rembrandt are categorized as Baroque style. The former painted altarpieces and introduced many innovations in painting such as dramatic lighting effects and the latter generated massive contributions during the Dutch Golden Age and influenced many young artists and new genres afterwards. 

The right middle side of figure \ref{fig:vit16-2} is Medieval style. Medieval period is thousand plus years between the division of the Roman Empire into Eastern and Western empires around the $4^{th}$ century AD and the beginnings of the Renaissance in Europe. (oxfordartonline) The medieval artists in this dataset are Duccio di Buoninsegna and Giotto di Bondone with styles of high-contrast, religious and represents many decorative objects such as stained glass, manuscripts, enamels, reliquaries and embroidered vestments.

On the bottom side of the figure \ref{fig:vit16-2}  is the dominant area by impressionism, which is a influential artistic movement arising in late $19^{th}$ century France. The artists from this period, such as Monet, Vincent van Gogh, Pissaro, rejected the system of state-controlled academies and salons in favor of independent exhibitions since 1874 and chose to capture the contemporary landscapes and nature scenes of model life, bourgeois leisure and recreation. The style of impressionism evolves over time and the some artists, however labeled as ``impressionism" , actually are really different from each other in terms of their own characteristic. So some impressionism artworks was closer to Romanticism and even Baroque. And the most distinguishable styles in the impressionism category is mostly from Claude Monet, which showed by on the very bottom of the scatterplot.

However, tSNE was only able to show clusters of groups with distinct characteristics, for Romanticism and Renaissance, they are kind of scattered around and are relatively difficult to trace.

 \textbf{LLE and Laplacian Eigenmap - Local and Global Connectivity} 
 
The visualization of LLE and Laplacian Eigenmap are similar. Given the nature that LLE and Laplacian tend to find global and local connectivity of each datapoint, they could form a ``flow", rather than a ``ball" created by tSNE. A ``flow" shape preserves better geometric distance different artworks given many of them are not created at the same period of time, and they have internal connections that cannot be easily described by clustering-liked visualizations. 

\begin{figure}[!ht]
    \centering
    \includegraphics[width=0.9\textwidth]{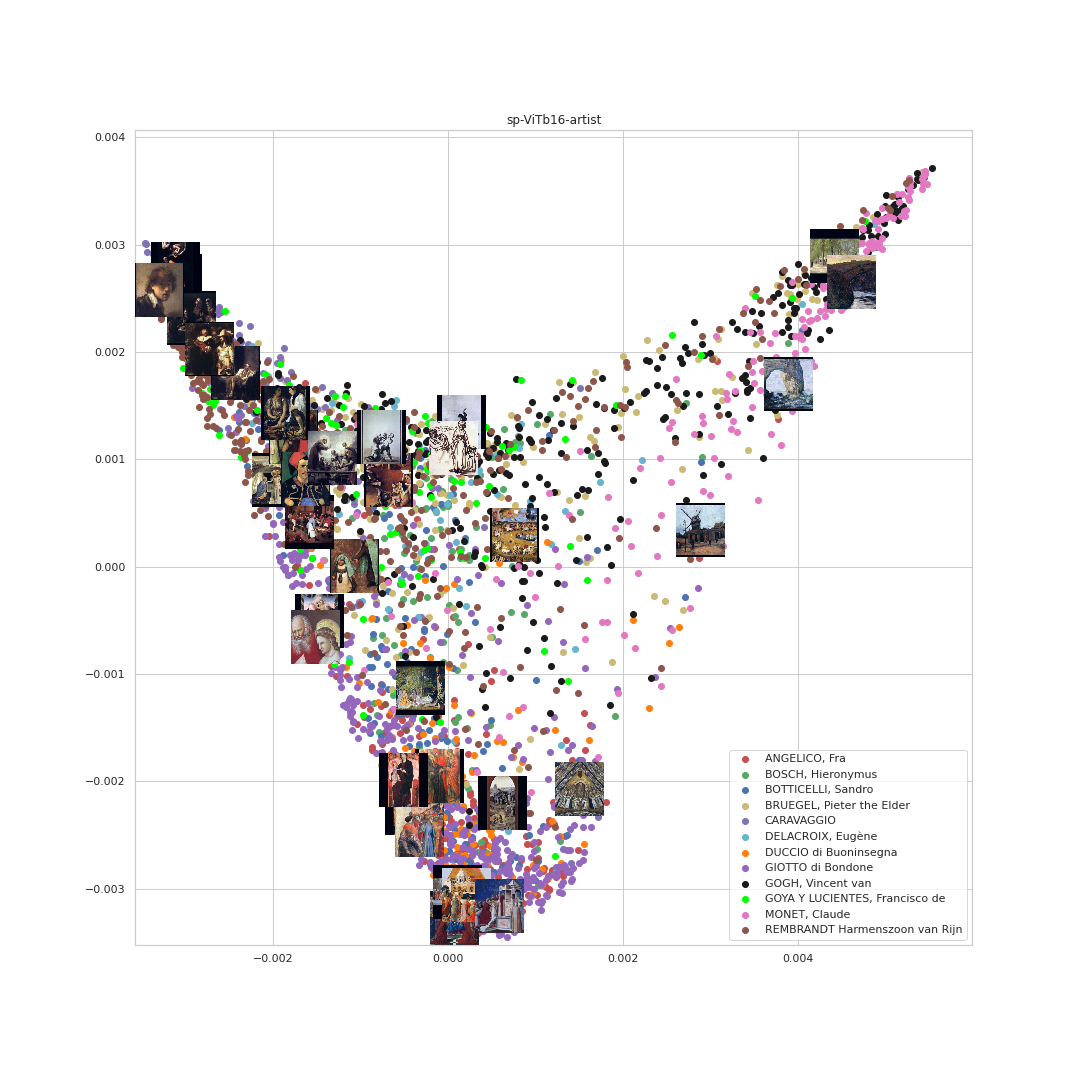}
    \caption{vit16-LapEig-artist}
    \label{fig:vit16-LapEig-artist}
\end{figure}

\begin{figure}[!ht]
    \centering
    \includegraphics[width=0.9\textwidth]{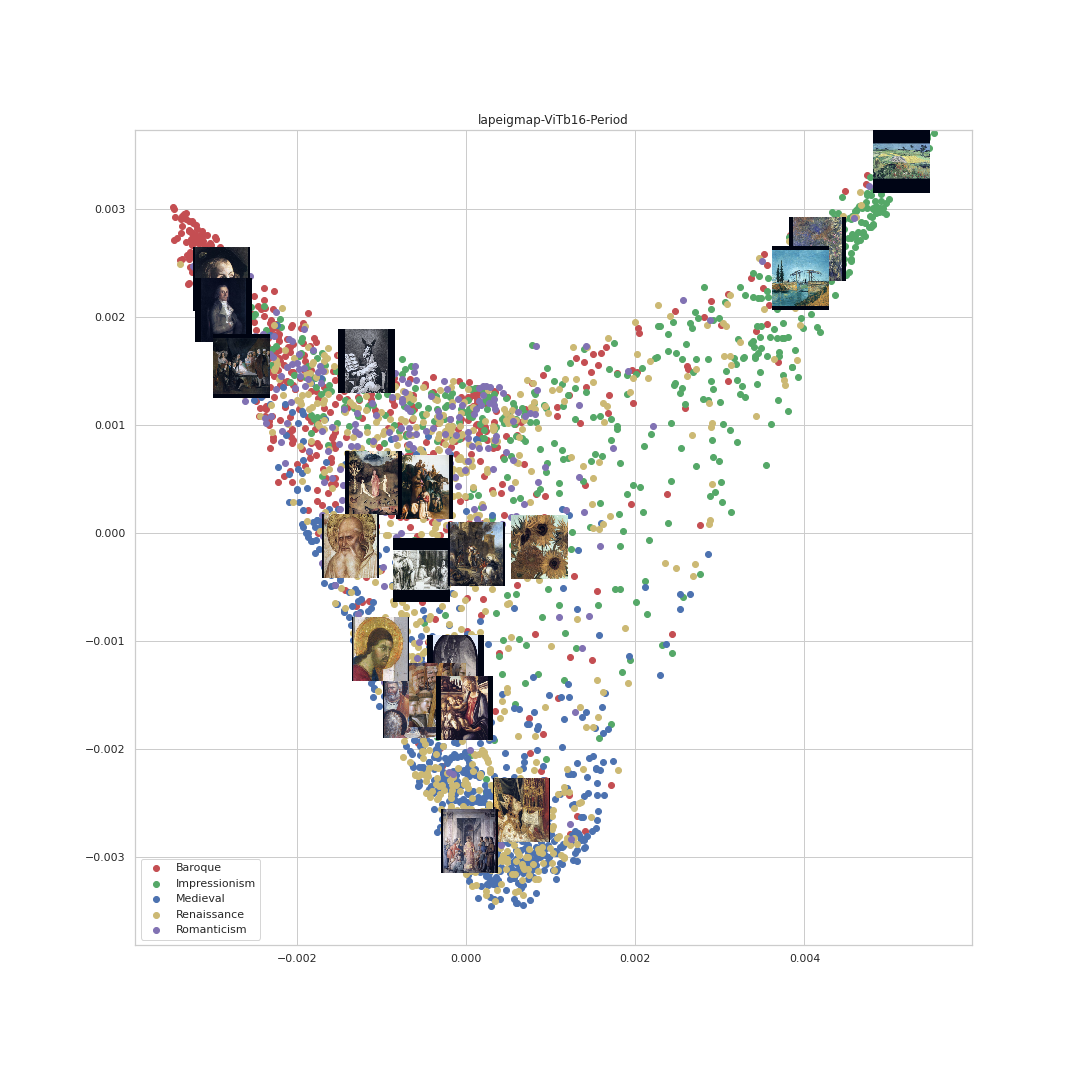}
    \caption{vit16-LapEig-period}
    \label{fig:vit16-LapEig-period}
\end{figure}

Take Figure \ref{fig:vit16-LapEig-period} and \ref{fig:vit16-LapEig-artist} as examples, starting from the bottom side, the blue dots represented the medieval period. Then the golden dots (Renaissance) spanned from bottom all the way to the top left and overlapped with both Medieval and Baroque. Renaissance period refers to the era of Europe from the $14^{th}$ to the $16^{th}$ century in which a new style in painting, sculpture and architecture developed after the Gothic. During Renaissance, many features of the medieval persisted, including the heritage of the artistic techniques used in books, manuscripts, precious objects and oil painting.(oxfordonline) So there are a large amount of overlap of blue and gold dots. 

The artists in the annotation dataset representing renaissance are Fra Angelico, Hieronymus Bosch from early Renaissance, Sandro Botticelli and Pieter Bruegel the Elder from Northern Renaissance. Despite the intrinsic differences of those two art style within the Renaissance larger umbrella, those two groups are merged as "Renaissance" in previous data preprocessing and transformation steps. 

Still looking at Figure \ref{fig:vit16-LapEig-period}, roughly between Baroque and Impressionism , there is a small group of purple dots representing Romanticism, by artists Eugene Delacroix and Francisco Jose de Goya y Lucientes. Romanticism is a movement in the arts and literature which originated in the late $18^{th}$ century, emphasizing inspiration, subjectivity, and the primacy of the individual and is a reaction against the order and restraint of classicism and neoclassicism. Romanticism does not have a very distinguishable style in this dataset as the sampled artworks are influenced by many other dominating art styles at the same time. This explains the reason why the dots of Romanticism are more scattered everywhere in the visualization. 

Finally, a group of Impressionism dots started from Baroque, spanned through Romanticism and formed its own distinct group of styles on the far right corner, which are contributed by Claude Monet and Vincent van Gogh. Since Claude Monet has a more consistent art style, his dots are located at relatively far corners, whereas Vincent van Gogh's early career artworks are relatively darker and looks more like baroque and late career artworks are closer to Impressionism. So the dots representing his artworks are more scattered than the ones from Claude Monet.

\begin{figure}[!ht]
    \centering
    \subfigure[ViT16-Laplacian Eigenmap-by artist]{\includegraphics[width=0.4\textwidth]{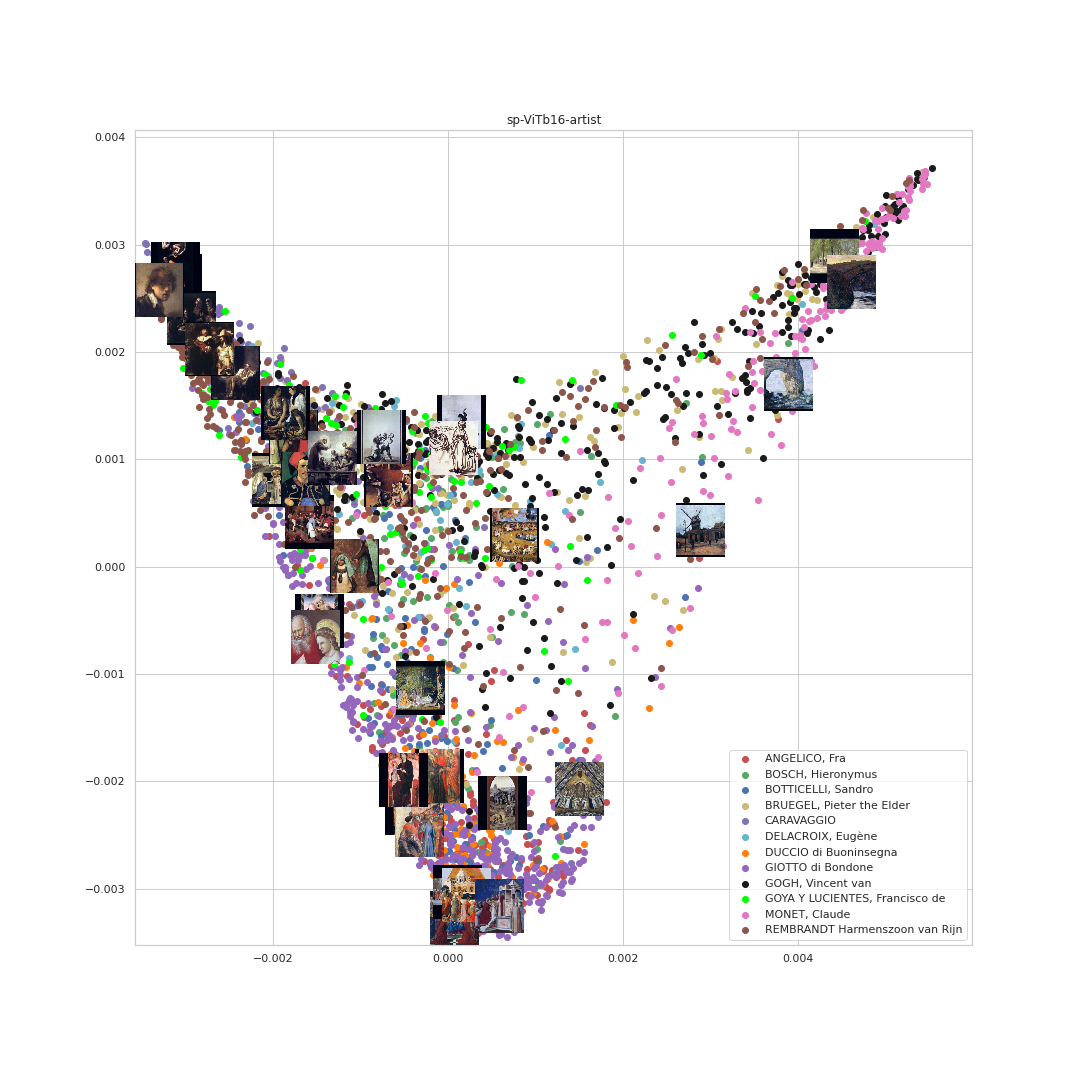}} 
    \subfigure[ViT16-Laplacian Eigenmap-by period]{\includegraphics[width=0.4\textwidth]{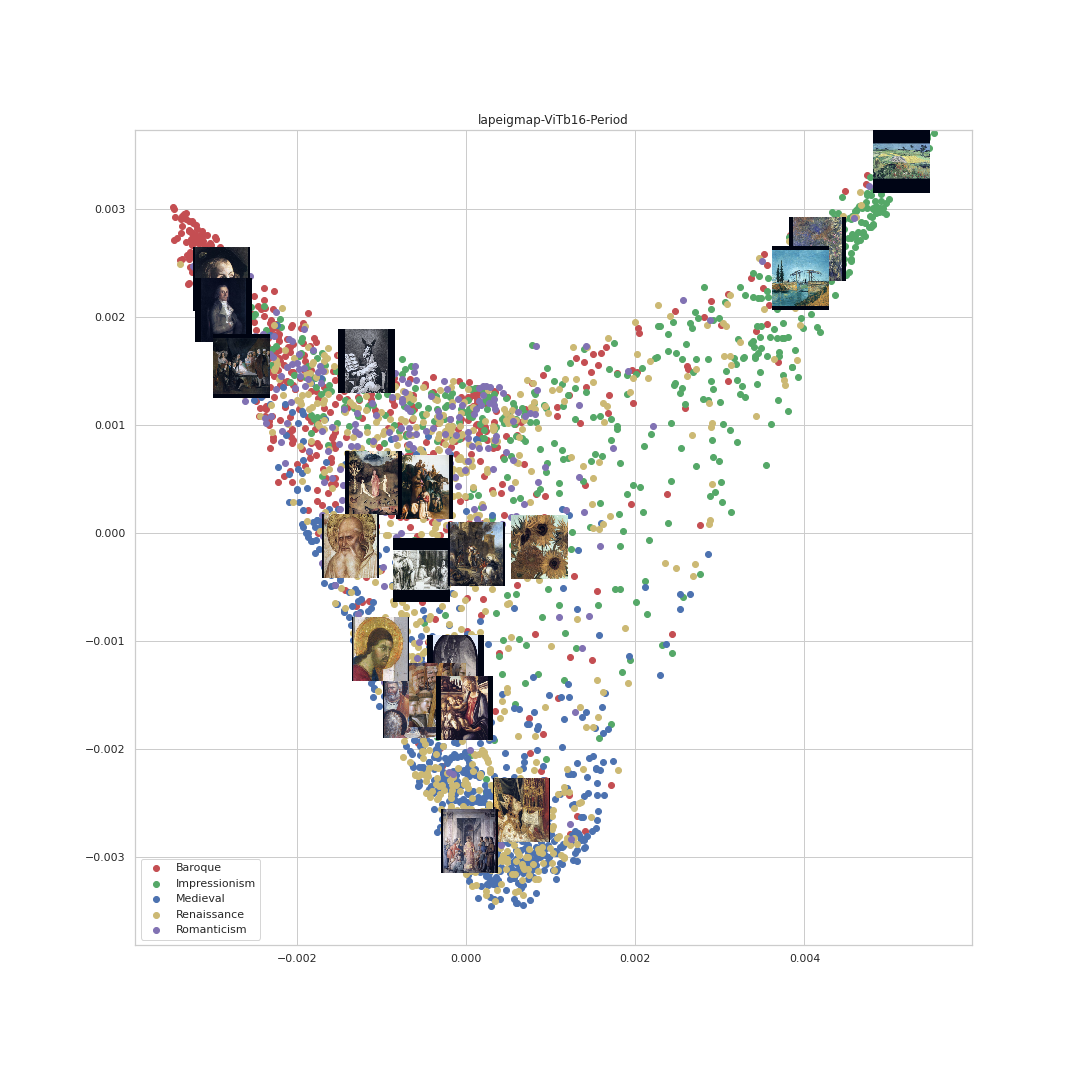}}
    \subfigure[ViT16-LLE-by Artist]{\includegraphics[width=0.4\textwidth]{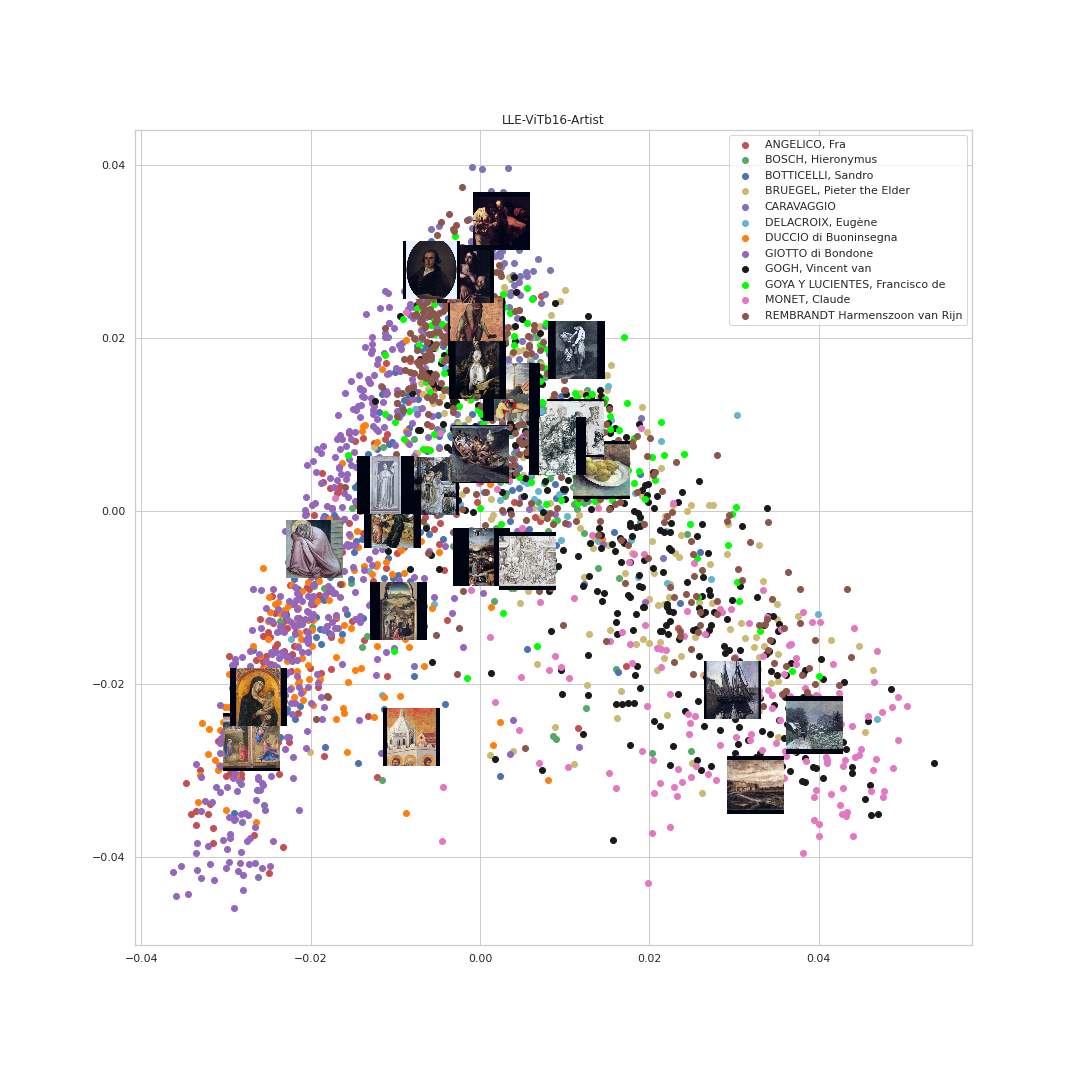}} 
    \subfigure[ViT16-LLE-by-Period]{\includegraphics[width=0.4\textwidth]{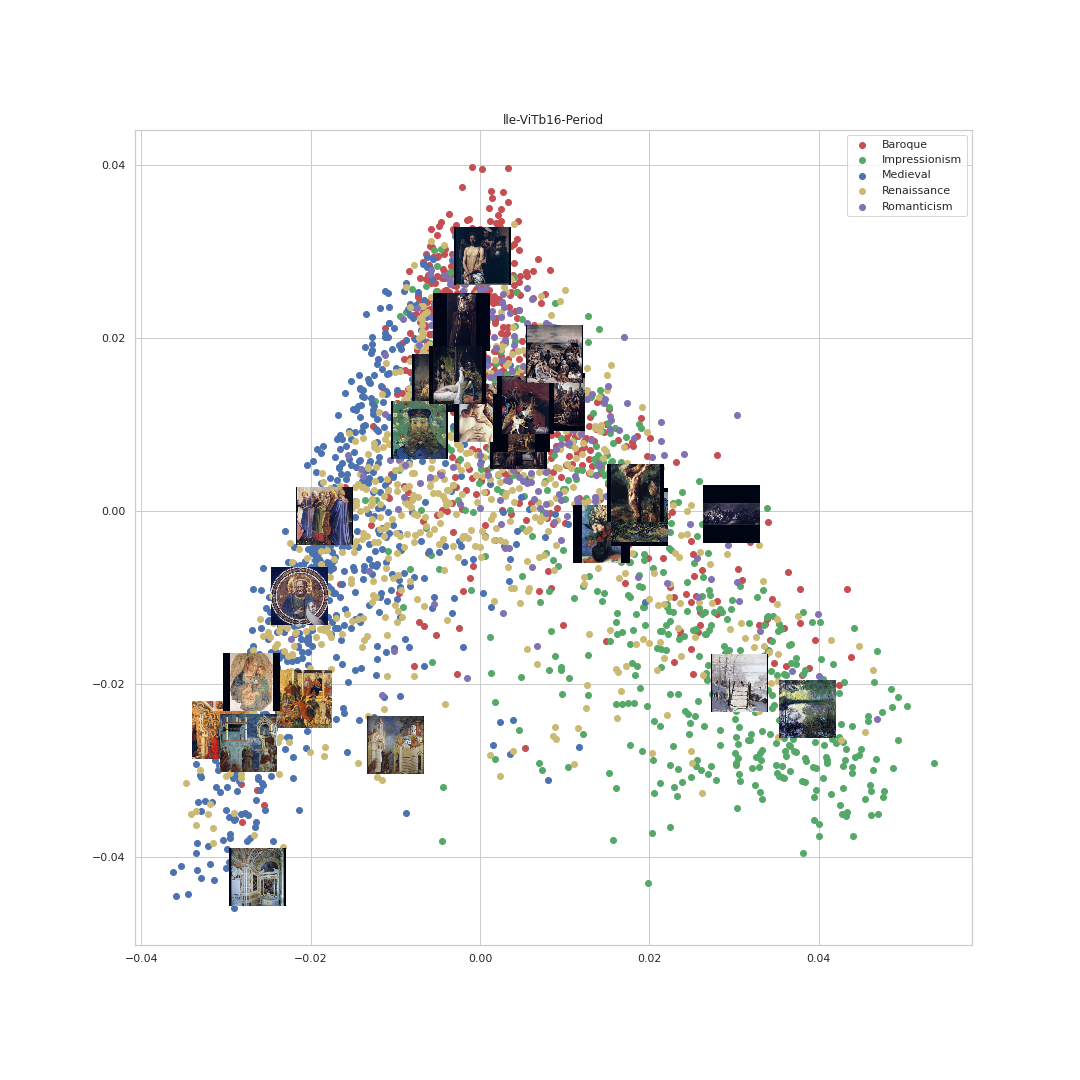}}

    \caption{Pre-trained ViT16 by Artists and Periods}
    \label{fig:vit16-3}
\end{figure}

\clearpage

\subsection{Training a Neural Network}

With all the pre-trained neural network feature embedding and manifold learning algorithm visualization results shown above, the winner seems obvious from the performance evaluation steps. Then a further question arise: Is a pre-trained neural network weights good enough? Can we improve a neural network's performance by performing further training of on top of the pre-trained model?

As mentioned previously, images and texts are completely different in terms of information structures. Unlike NLP, which we can leverage current state-of-the-art focuses on using BERT models for pre-training (D.Jacobs et al., 2019) that incorporate large text corpora on general subjects such as Wikipedia and Toronto books will achieve a good starting poi(S. Humeau et al., 2019), the pre-training of image data is very task-sensitive and the pre-trained parameters might not be a good fit. The images used to pre-train the neural network models for Vision Transformers are mainly cats, dogs, goldfish, etc, which are not even close to the art images used in this task.

In this project, the hypothesis is that if the neural network uses specific artwork data and annotation file for training and then perform the feature embedding after ``seeing" those images , the feature embedding results can be improved. Then, with a better feature embedding, the manifold learning visualization results would then show more clear patterns regarding different classes of styles of art works and the trends of how art history evolves.

To implement the training steps, the image dataset is split into train and validation in a rate of 0.8:0.2, no testing data is used here since this project is only interested at in-sample performance improvements. Since it is often beneficial to fine-tune at higher resolution than pre-training (H. Touvron et al., 2019; A. Kolesnikov et al., 2020), original pictures downloaded from website are used for training without any compression.  

In order to adapt the structure of training data, the classification head of ViT16 model is modified and the number of output features are changed from default 1000 to number of classes. Since the classification head is implemented by a MLP with one hidden layer at pre-training time and by a single linear layer at fine-tuning time (A. Dosovitskiy et al, 2021), the single linear layer in the classification head was taken out for feature embedding after training completed. 

\begin{figure}[!ht]
    \centering
    \includegraphics[width=0.9\textwidth]{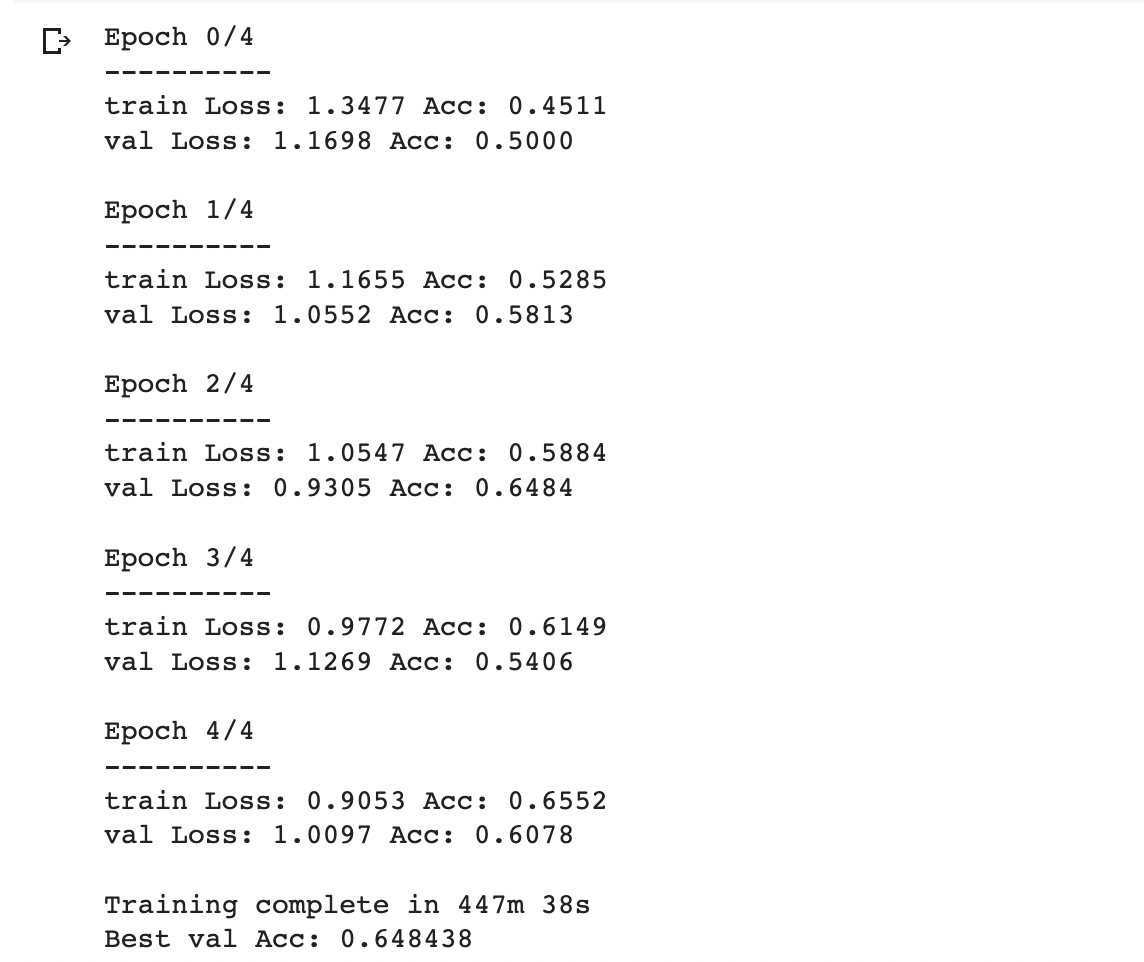}
    \caption{Training Logs ViT16}
    \label{fig:trainvit16}
\end{figure}

As shown in the table ans also in \ref{fig:trainvit16}, the model is trained at Google Colab using GPU. The training is a long process given the infrastructure constraints so the number of epoch is reduced from originally planned 20 to 5. The model is trained for 5 full epochs using stochastic gradient descend optimizer with batch size = 5. 

\begin{tabular}{|c|c|c|c|c|}
\hline
\textbf{epoch} & \textbf{train loss} & \textbf{train acc} & \textbf{val loss} & \textbf{val acc}\\\hline
    0 & 1.3477& 0.4511 & 1.1698 & 0.5000\\\hline
    1 & 1.1655 & 0.5285 & 1.0552 & 0.5813 \\\hline
    2 & 1.0547 & 0.5884 & 0.9305 & 0.6484 \\\hline
    3 & 0.9772 & 0.6149 & 1.1269 & 0.5406 \\\hline
    4 & 0.9503 & 0.6552 & 1.0097 & 0.6078\\\hline
\end{tabular}

On average, each epoch takes about 1.5-2 hours to train, the full training took roughly 7.5 hours to finish. The validation accuracy started at 0.5 and then gradually increased and the highest validation accuracy is 0.645, stating the training is effective and the neural network is learning from the new training data on top of pre-trained knowledge. The training is pretty effective given the progresses made with limited time of training epochs, and the model is likely to achieve better validation accuracy if there are more training epochs performed.

\subsection{Evaluating Training Results}

To analyze the result of training a neural network more visually, similar manifold learning and visualization methods are applied to the feature embedding generated by the trained ViT16 model to see if the visualization would give a different look. 

As mentioned previously, the single linear layer in the classification head was taken out for feature embedding after training is completed, which has a total of 768 of feature outputs, slightly different from the output number of features of a pre-trained ViT16 model (1000) but no fundamental differences when passed into manifold learning algorithms to learn the geometric structure at high-dimension and visualize in a low-dimension embedding. 

\begin{figure}[!ht]
    \centering
    \subfigure[Trained-ViT16-LaplacianEigenmap]{\includegraphics[width=0.45\textwidth]{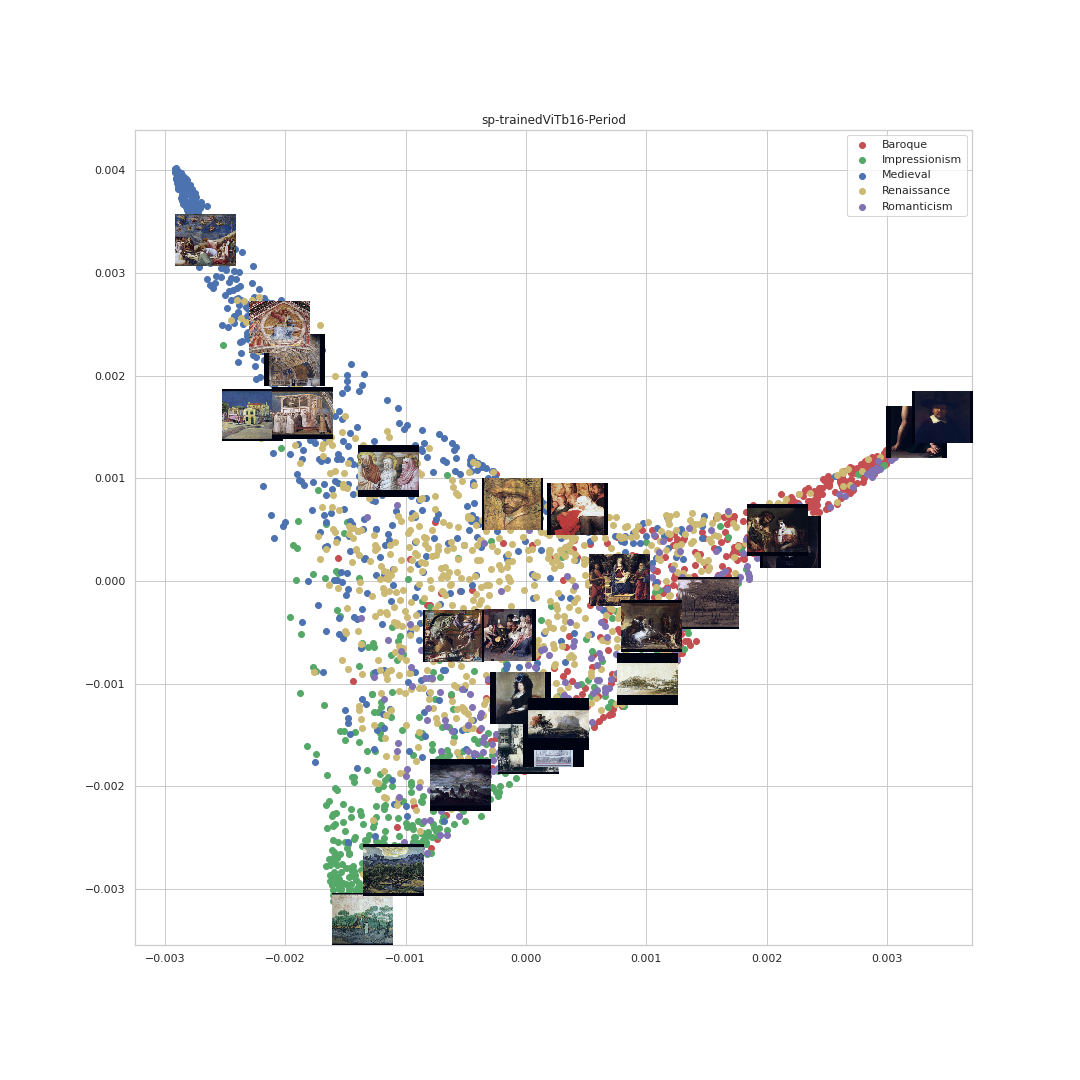}} 
    \subfigure[Trained-ViT16-tSNE]{\includegraphics[width=0.45\textwidth]{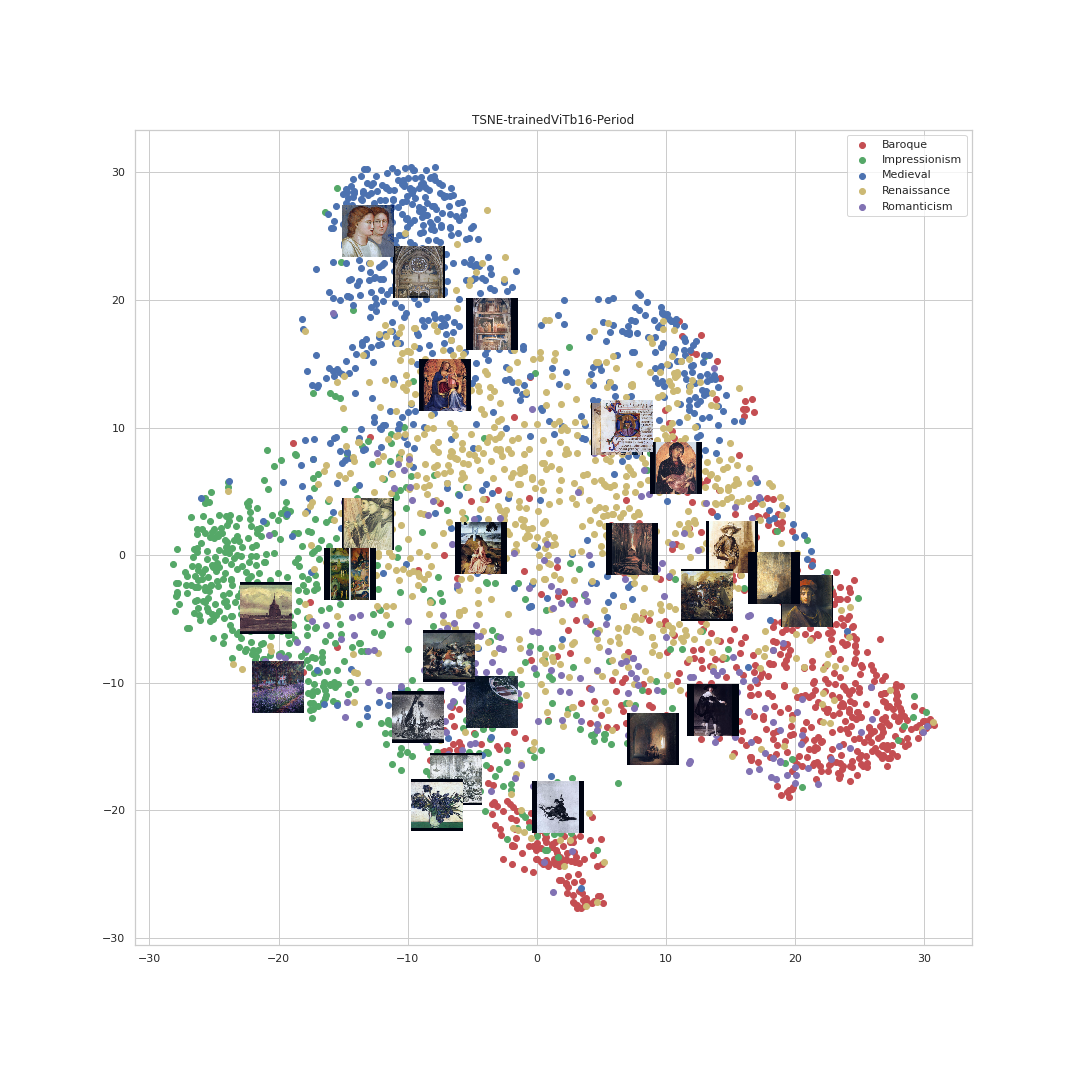}}
    \subfigure[Trained-ViT16-LLE]{\includegraphics[width=0.45\textwidth]{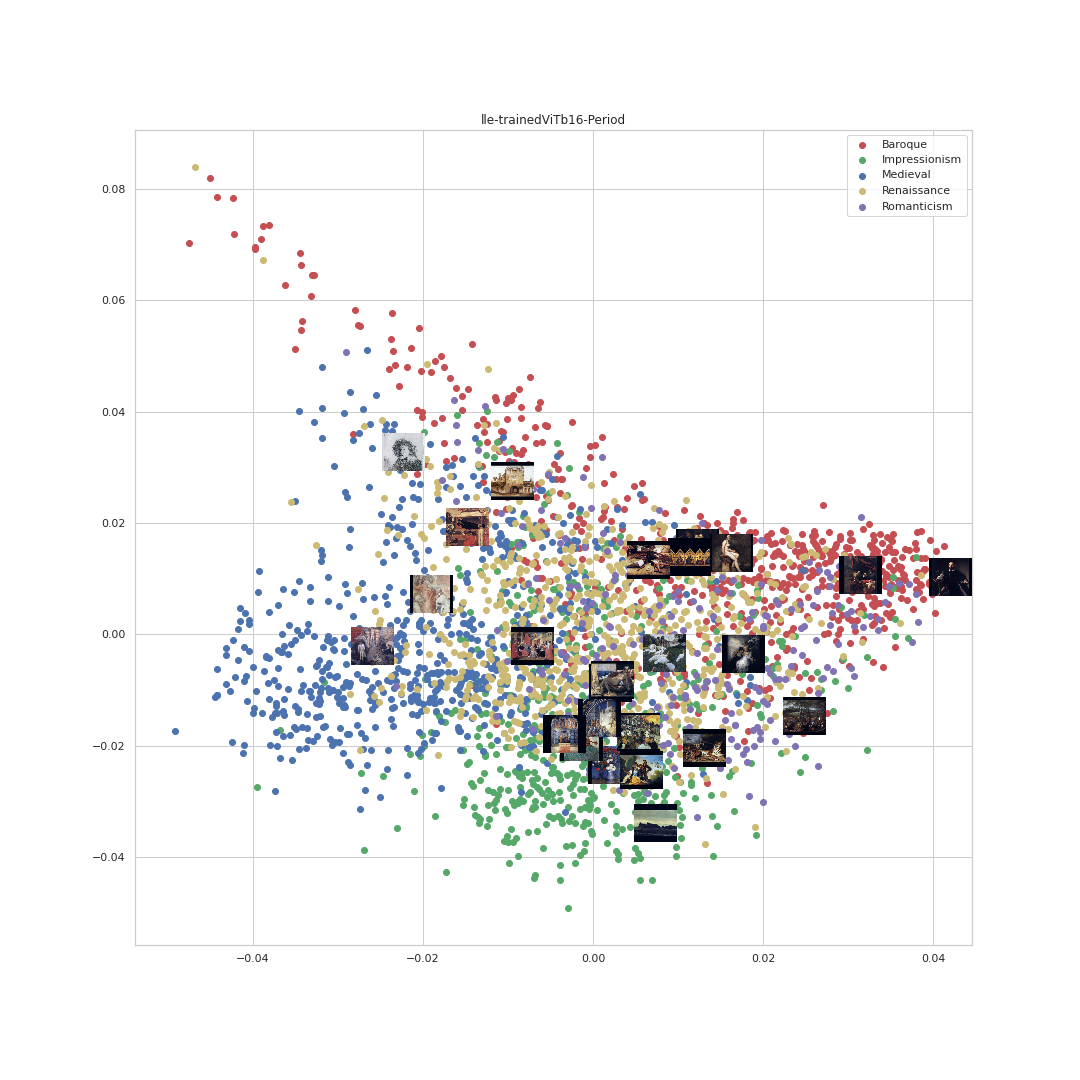}} 
    \subfigure[Trained-ViT16-ISOMAP]{\includegraphics[width=0.45\textwidth]{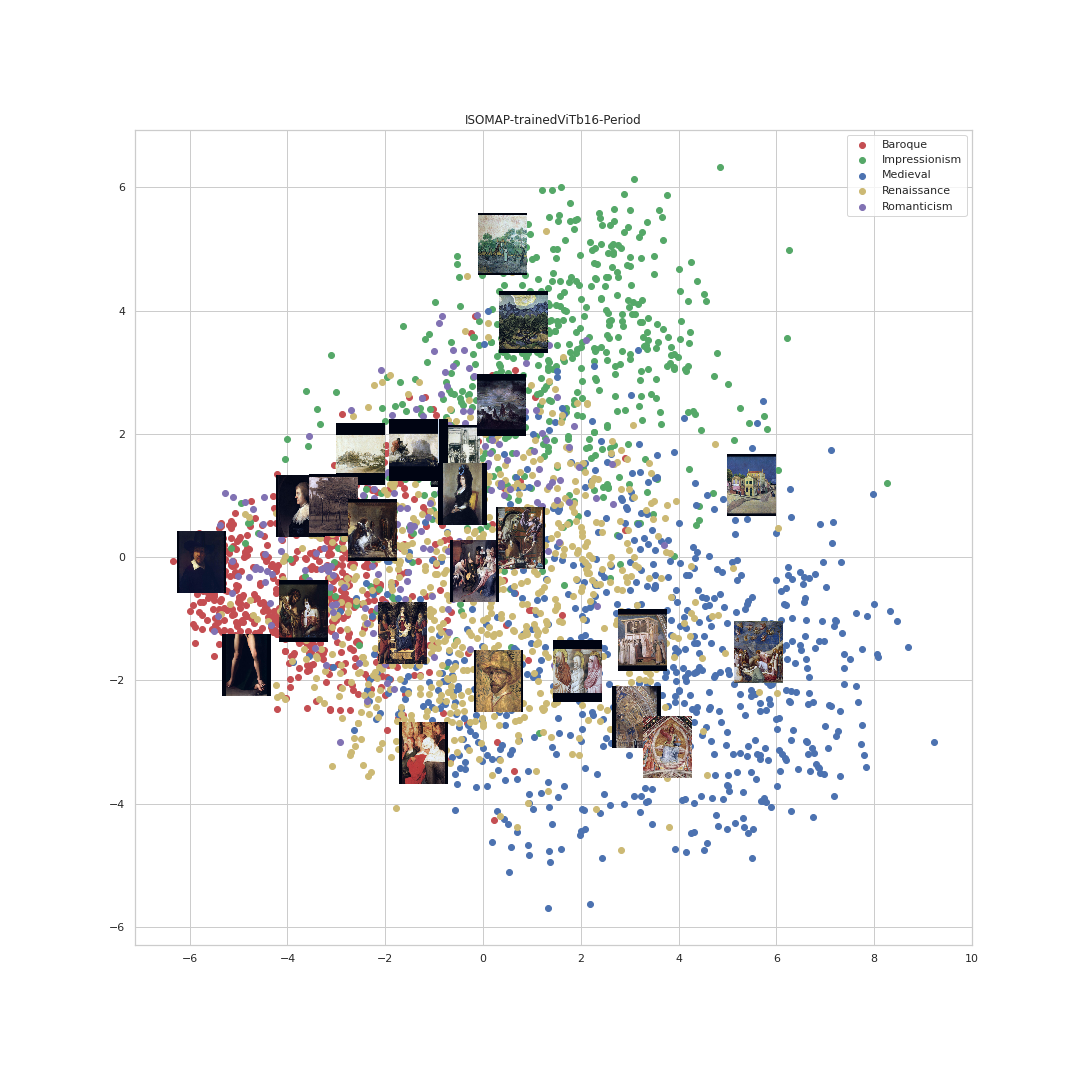}} 

    \caption{Feature Embedding Visualization After Training}
    \label{fig:vit16-trained}
\end{figure}

LLE, ISOMAP, Spectual Clustering (Laplacian Eigenmap) and tSNE are applied to the feature embedding of post-trained ViT16 model. From the visualizations shown in Figure \ref{fig:vit16-trained}, there is a clear improvement across all selected manifold learning algorithms for visualizing the feature embedding. There are more distinct separation of different art categories over the time, and the pattern showing the connectivity among different classes are more obvious, For example, the shape of LLE is significantly different pre-post training, and it is able to see Renaissance as a center of inspiration that inherited or nurtured all other art categories while maintain its own distinct style. Similar pattern are also captured by all other manifold learning algorithms. 

Taking a closer look at Laplacian Eigenmap (Spectral Clustering), comparing to the previous version, the overlaps of points with different colors reduced significantly, especially the Renaissance class is now more centered at the graph and has close connections to almost every art style, versus the previous version having a large portion overlaying on top of medieval. 

There is also a huge improvement of tSNE. For feature embedding done by pre-trained models, despite the ability of finding rough clusters of different categories, tSNE still treated the entire dataset as a ``big ball", meaning it was not able to find significant distance between artworks from different categories. With post-trained feature embedding, however, tSNE formed a star-shaped structure and each "branch" represents a distinct art style such as Impressionism, Medieval, Baroque. Given Romanticism has less datapoint and the style is more like a fusion, so it's kind of scattered around. Renaissance revolutionized art history and had deep and profound impact on different art styles, so the cluster is at the center of the data cloud and has some overlaps to almost each of the categories. 

Similar to the pre-trained version, several sub-clusters are generated by tSNE with post-trained feature embedding, yet the difference are more distinct for the datapoints actually belong the same category, meaning the algorithm learned from the training session and was able to correctly classify the artworks with slight style difference as the same group. Yet there are still some artworks looks very unconventional given the categories they belong to. For example, Baroque has a smaller sub-cluster that is closer to impressionism, and Medieval has a smaller sub-cluster that has more overlap with renaissance. Those interesting observation could be mainly due to some artists, like Rembrandt and Vincent Van Gogh, have more than one distinct style across their career time. 

\begin{figure}[!ht]
    \centering
    \includegraphics[width=0.9\textwidth]{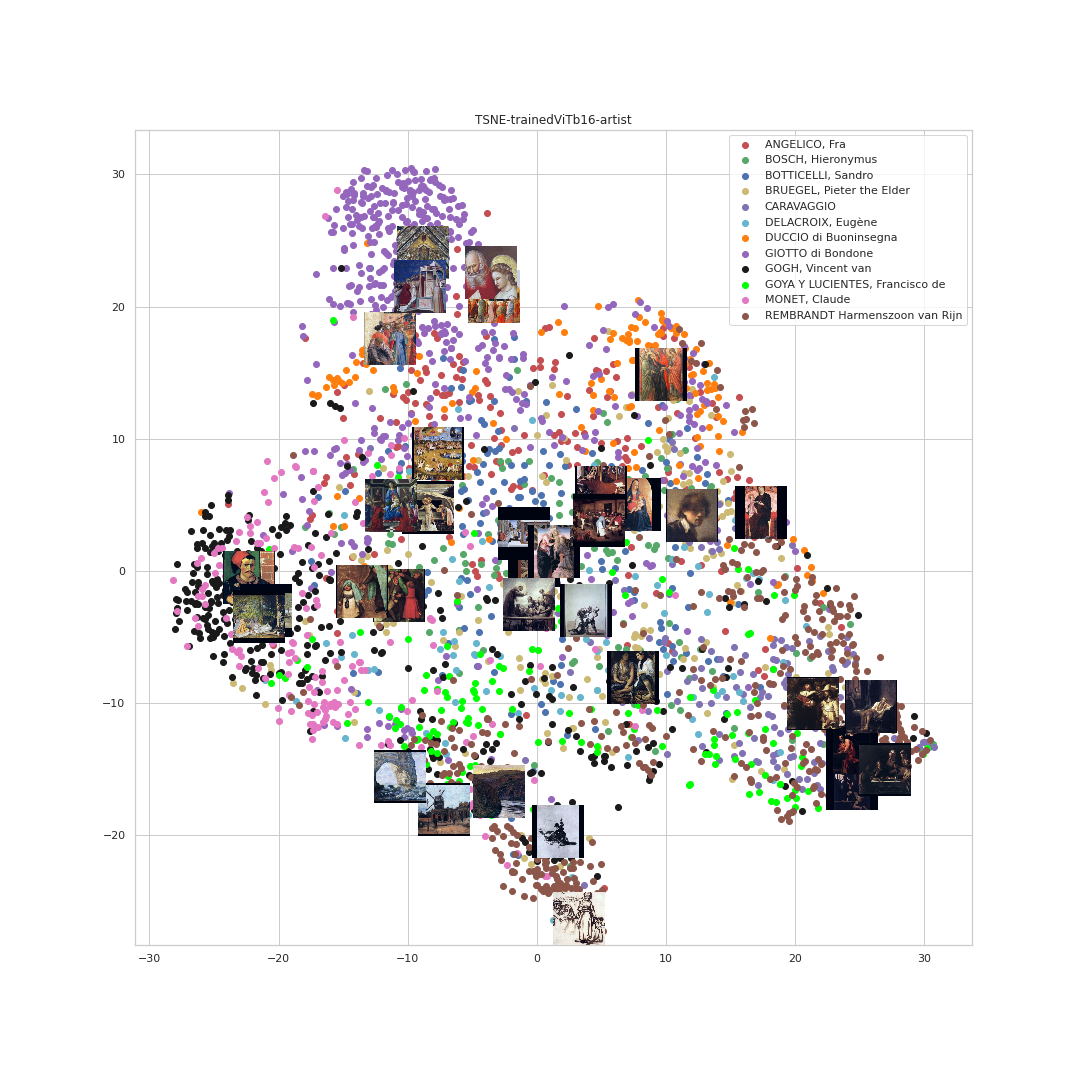}
    \caption{trained-vit16-tSNE-artist}
    \label{fig:tsne-trained-vit16-artist}
\end{figure}

Referring to Figure \ref{fig:tsne-trained-vit16-artist}, one can tell that Rembrandt has a very distinct Dutch baroque style, but he also draws a lot of sketches, which are considered as a standard-along cluster because those artworks are very different from a typical baroque style. Vincent Van Gogh's early career artwork are mainly very dark and looks close to baroque, later in his life he moved to Paris where he met Monet, Renoir and other impressionism artists hence his work was greatly influenced and look a lot more like impressionism.

\clearpage

\subsection{One Step Further: PHATE}
PHATE algorithm came into my attention in a relative late stage of this project yet it outperformed all other manifold learning algorithms illustrated before, hence I think it is worthwhile to discuss it here. 

PHATE (Potential of Heat-diffusion for Affinity-based Trajectory Embedding) is a powerful tool for visualizing high dimensional data which uses a novel conceptual framework for learning and visualizing the manifold to preserve both local and global distances. It has a wide range of applications that can be applied to datasets such as facial images and single-cell data from human embryonic stem cells(Moon, K.R et al., 2019) and financial data (R. Ding, 2023).

\begin{figure}[!ht]
    \centering
    \subfigure[Trained-ViT16-LaplacianEigenmap]{\includegraphics[width=0.45\textwidth]{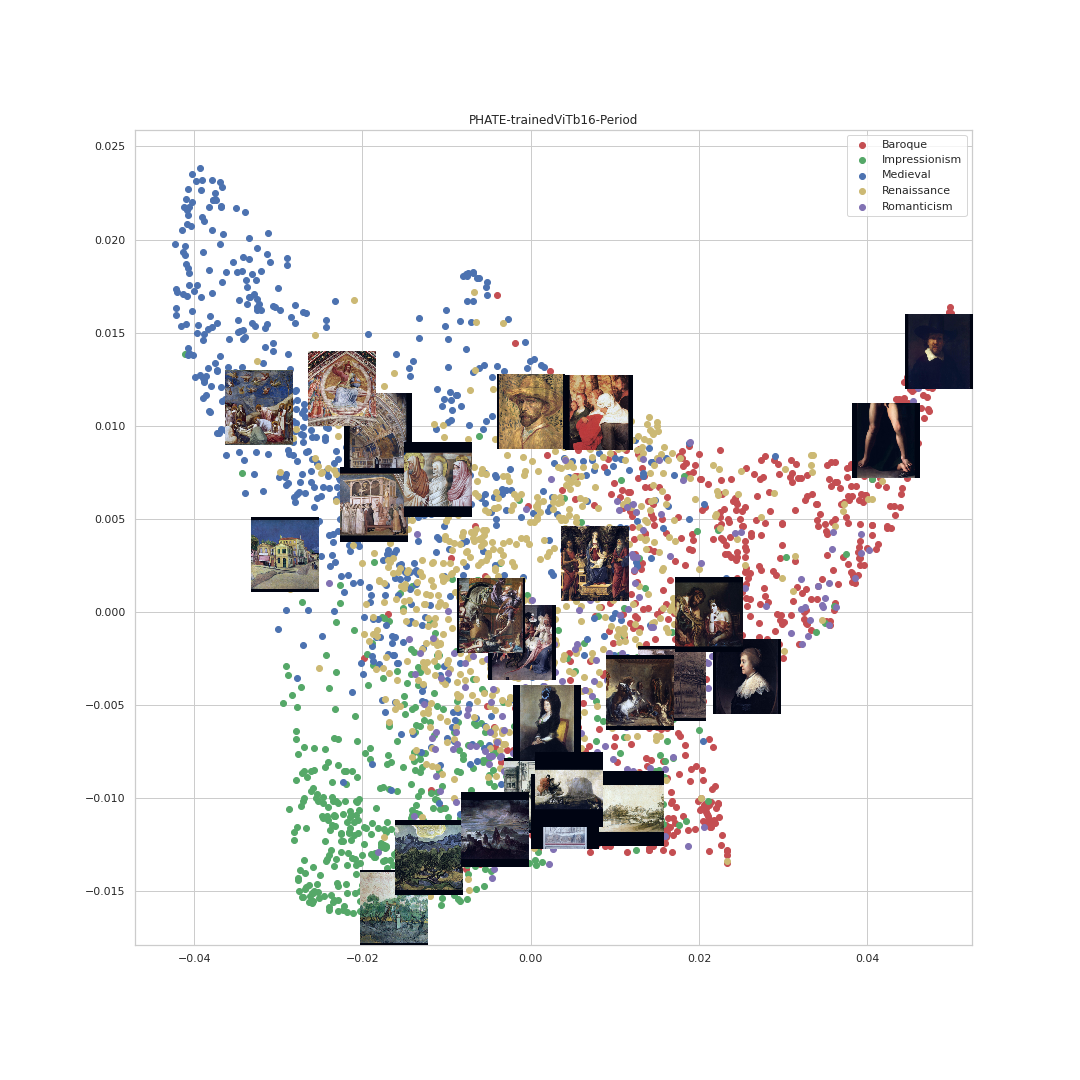}} 
    \subfigure[Trained-ViT16-tSNE]{\includegraphics[width=0.45\textwidth]{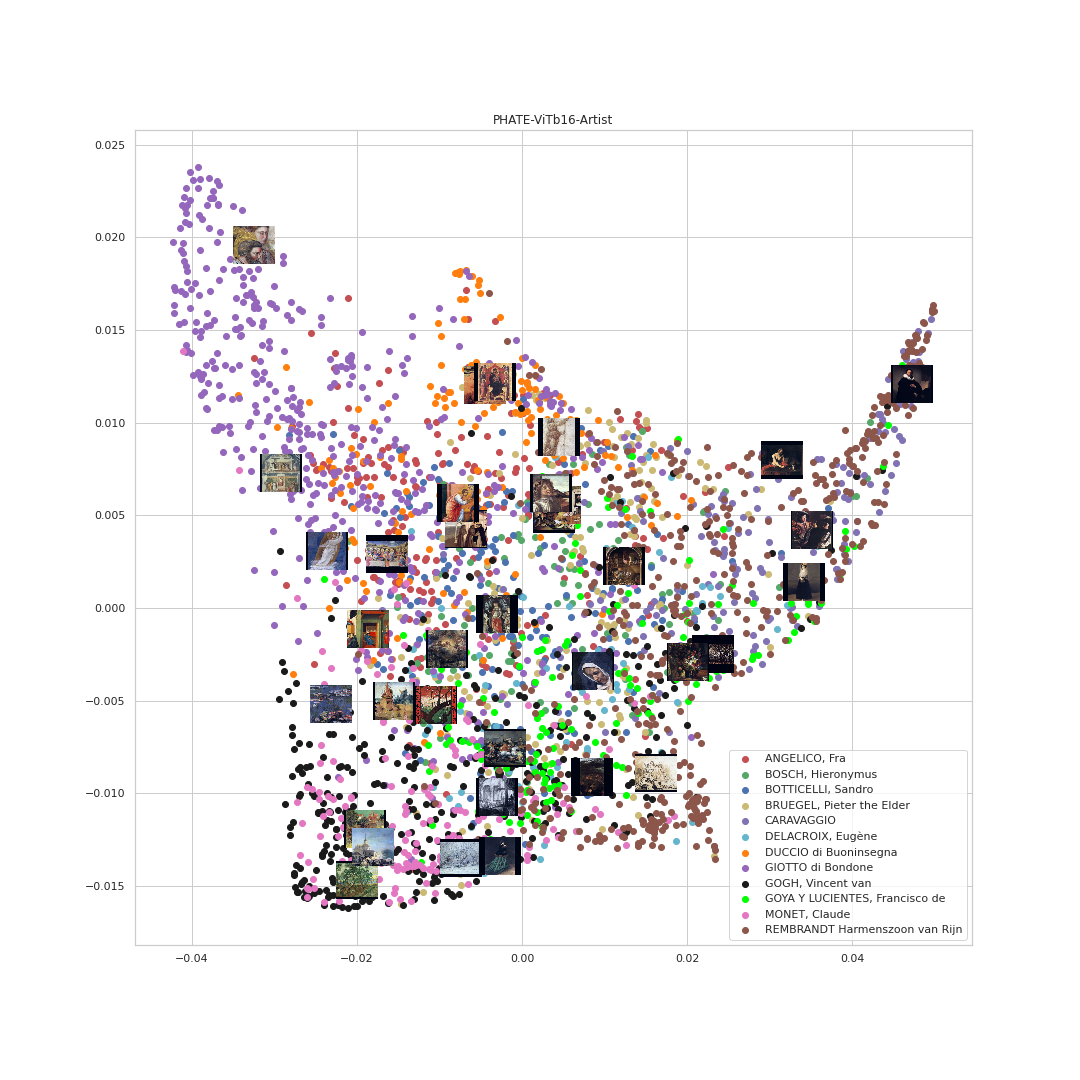}}

    \caption{Vit16 After Training - PHATE}
    \label{fig:phate-vit16-trained}
\end{figure}

Similar as implemented before, PHATE algorithm is applied to the feature embedding generated by the post-trained ViT16 model and visualized the results in a low dimension. As shown in Figure \ref{fig:phate-vit16-trained}, the PHATE algorithm shows a similar ``clustering and branching" power as tSNE, but with a more robust performance: Baroque, Impressionism and Medieval still dominates three different main directions, Renaissance is at the center and connects to each categories and Romanticism is scattered around, yet there is less overlap of Renaissance to different artworks comparing to tSNE, meaning that PHATE developed a better understanding that Renaissance is not only a transitional art style adjacent to other styles in different degrees, but also as a standalone category has its own style. 

Now we pivot to see how the art works from different artists are presented, (larger picture shown in appendix Figure \ref{fig:phate-trained-vit16-artist}) we can see that for Giotto and Duccio, although both labeled as Medieval artist, Giotto's artworks spans from typical Medieval to Renaissance, marking his clear career transition; on contrary, Duccio's style is relatively consistent. Fra Angelico and Sandro Botticelli are both early renaissance artists, so their artworks have plenty of overlap between Medieval and Renaissance. Eugene Delacroix is a Romanticism artist but his art style looks very close to baroque because of the high contrast of dark and light. And as previously observed already by tSNE, Vincent van Gogh, as an Impressionism artist, his artwork style looks closer to Baroque in his early career versus closer to impressionism later, so the plots representing his artwork appeared in different clusters. Same thing also happens to Rembrandt, as in this dataset he produced both typical baroque style art works as well as sketches. 

\clearpage 

\section{Conclusion}

After the exploration of mainstream state-of-art neural network models, ViT16 appears to produce the best feature embedding results, and among all the manifold learning algorithms, PHATE gave the best visualization in a low dimension embedding, tSNE is also doing a good job in visualization. Laplacian Eigenmap and LLE preserved the best geometric structure of the manifold in its original high dimension feature in their globally or locally connectivity approach. 

``The Shape of Art History in the Eyes of the Machine" was published in 2018, while AlexNet, VGGNet and ResNet, as the state-of-art neural network models, were chosen to implement and analyze. Both Vision Transformer and PHATE were still in making - Vision Transformer came out in 2019, then PHATE in 2020. This project successfully captured such important timing by comparing a subset of earlier state-of-art neural network models and classic manifold learning algorithms as well as Vision Transformer and PHATE. It turned out that Vision Transformer is the best performing neural network model in this particular task and the performance is even better after a 5-epoch short training session, and PHATE was proved to be the top (if not the best) manifold learning and visualization algorithm for the feature embedding. 

Due to infrastructure constraints, this project only used a very small size of art history data and the training process is relatively short and lightweight mainly just for illustrating the methodology. If such approach were implemented at a larger scale as in the original paper, there could be even more interesting findings looking into the shape of art history through the eyes of the new machine.

\section{Acknowledgements}
This project is inspired by ``The Shape of Art History in the Eyes of the Machine", thanks to Elgammal, Ahmed and team for pioneering the research and implementation. Thanks to Rui Ding for project idealization and introducing PHATE algorithm. Thanks to all the authors and team for designing and implementing neural networks and manifold learning algorithms explored in this project. Thanks to Xian Yang for useful discussions in art history domain. 

\section{Reference}

\begin{enumerate}
  \item Kingma, Diederik \& Ba, Jimmy. (2014). Adam: A Method for Stochastic Optimization. International Conference on Learning Representations. 
  
  \item K. He, X. Zhang, S. Ren and J. Sun, "Deep Residual Learning for Image Recognition," 2016 IEEE Conference on Computer Vision and Pattern Recognition (CVPR), 2016, pp. 770-778, doi: 10.1109/CVPR.2016.90.
  
  \item Krizhevsky, Alex \& Sutskever, Ilya \& Hinton, Geoffrey. (2012). ImageNet Classification with Deep Convolutional Neural Networks. Neural Information Processing Systems. 25. 10.1145/3065386. 
  
  \item Tenenbaum, Joshua \& Silva, Vin \& Langford, John. (2000). A Global Geometric Framework for Nonlinear Dimensionality Reduction. Science. 290. 2319-2323. 
  
  \item M. Belkin and P. Niyogi, "Laplacian Eigenmaps for Dimensionality Reduction and Data Representation," in Neural Computation, vol. 15, no. 6, pp. 1373-1396, 1 June 2003, doi: 10.1162/089976603321780317.

  \item Dosovitskiy, Alexey \& Beyer, Lucas \& Kolesnikov, Alexander \& Weissenborn, Dirk \& Zhai, Xiaohua \& Unterthiner, Thomas \& Dehghani, Mostafa \& Minderer, Matthias \& Heigold, Georg \& Gelly, Sylvain \& Uszkoreit, Jakob \& Houlsby, Neil. (2020). An Image is Worth 16x16 Words: Transformers for Image Recognition at Scale. 
  
  \item Donoho, David \& Grimes, Carrie. (2003). Hessian eigenmaps: Locally linear embedding techniques for high-dimensional data. Proc. National Academy of Science (PNAS), 100, 5591-5596. Proceedings of the National Academy of Sciences of the United States of America. 100. 5591-6. 10.1073/pnas.1031596100. 
  
  \item Hout, Michael \& Goldinger, Stephen \& Brady, Kyle. (2014). MM-MDS: A Multidimensional Scaling Database with Similarity Ratings for 240 Object Categories from the Massive Memory Picture Database. PloS one. 9. e112644. 10.1371/journal.pone.0112644. 
  
  \item van der Maaten, Laurens \& Hinton, Geoffrey. (2008). Viualizing data using t-SNE. Journal of Machine Learning Research. 9. 2579-2605. 
  
  \item Hinton, Geoffrey E and Roweis, Sam. (2002). Stochastic Neighbor Embedding, Advances in Neural Information Processing Systems, MIT Press vol. 15. 
  
  \item Lee, Seolhwa \& Sedoc, João. (2020). Using the Poly-encoder for a COVID-19 Question Answering System. 10.18653/v1/2020.nlpcovid19-2.33. 
  
  \item Humeau Humeau, S., Shuster, K., Lachaux, M., \& Weston, J. (2019). Poly-encoders: Transformer Architectures and Pre-training Strategies for Fast and Accurate Multi-sentence Scoring. arXiv: Computation and Language.
  
  \item Elgammal, Ahmed \& Mazzone, Marian \& Liu, Bingchen \& Kim, Diana \& Elhoseiny, Mohamed. (2018). The Shape of Art History in the Eyes of the Machine. 
  
  \item Hugo Touvron, Andrea Vedaldi, Matthijs Douze, and Herve Jegou. (2020). Fixing the train-test resolution discrepancy: Fixefficientnet. arXiv preprint arXiv:2003.08237.
  
  \item Moon, K.R., van Dijk, D., Wang, Z., et al.
 Visualizing structure and transitions in high-dimensional biological data. (2019). Nat. Biotechnol. 37, 1482–1492, https://doi.org/10.1038/s41587-019-0336-3

\item Ding, R. Visualizing Structures in Financial Time Series Datasets through an Affinity-based Diffusion Transition Embedding. (2023). To appear in the Journal of Financial Data Science, Vol 5, Issue 1, Winter 2023.

\item Alexander Kolesnikov, Lucas Beyer, Xiaohua Zhai, Joan Puigcerver, Jessica Yung, Sylvain Gelly, and Neil Houlsby. Big transfer (BiT): General visual representation learning. In ECCV, 2020.

\item Devlin, Jacob, Chang, Ming-Wei, Lee, Kenton and Toutanova, Kristina", BERT : Pre-training of Deep Bidirectional Transformers for Language Understanding, Association for Computational Linguistics, 4171--4186, 2019

  \item https://www.oxfordartonline.com
  
  \item https://towardsdatascience.com/auto-encoder-what-is-it-and-what-is-it-used-for-part-1-3e5c6f017726 
  
  \item https://www.kaggle.com/code/krsnewwave/model-european-art-resnet-18/data
  
\end{enumerate}

\clearpage

\section{Appendix}

\begin{figure}[!ht]
    \centering
    \includegraphics[width=0.9\textwidth]{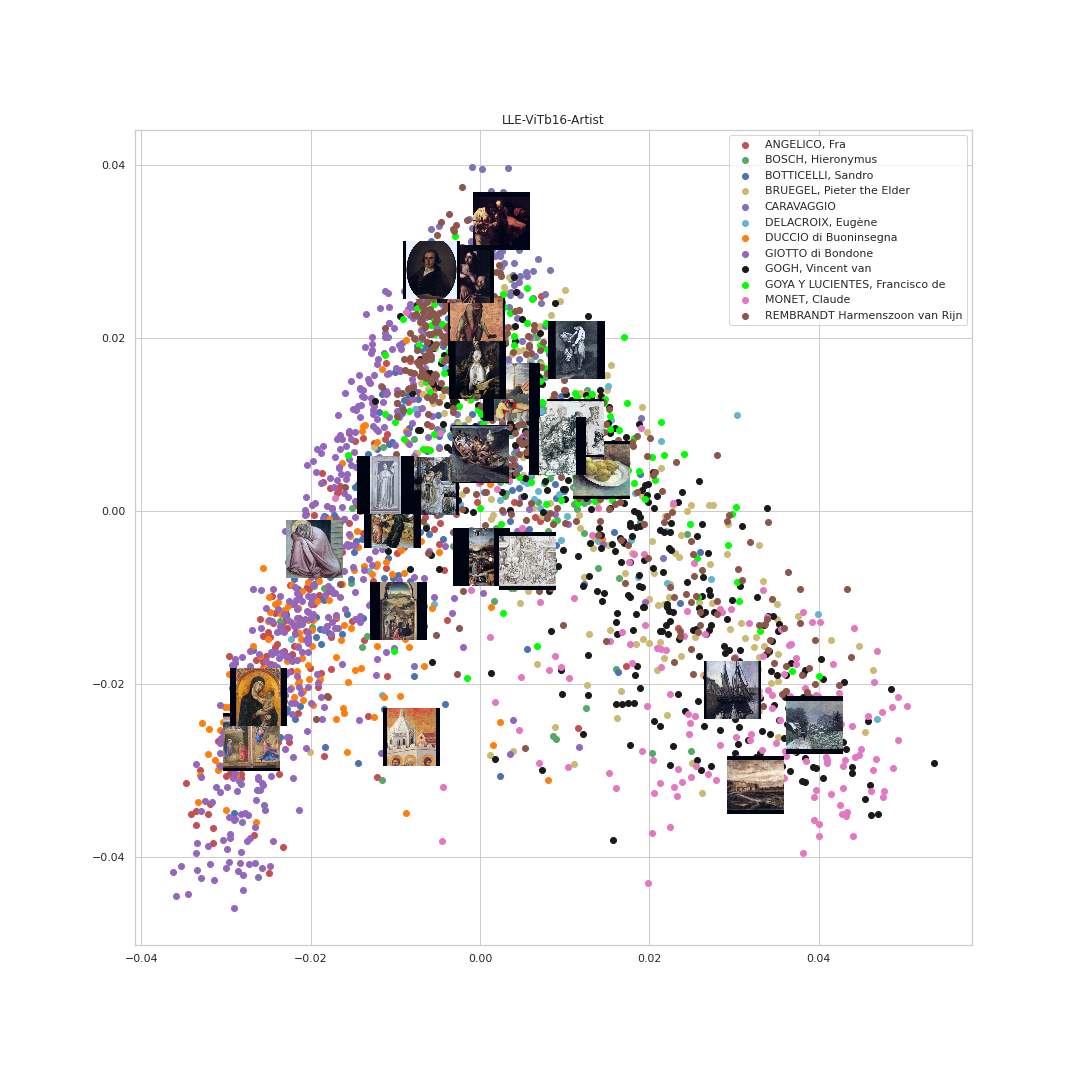}
    \caption{vit16-lle-artist}
    \label{fig:vit16-lle-artist}
\end{figure}

\begin{figure}[!ht]
    \centering
    \includegraphics[width=0.9\textwidth]{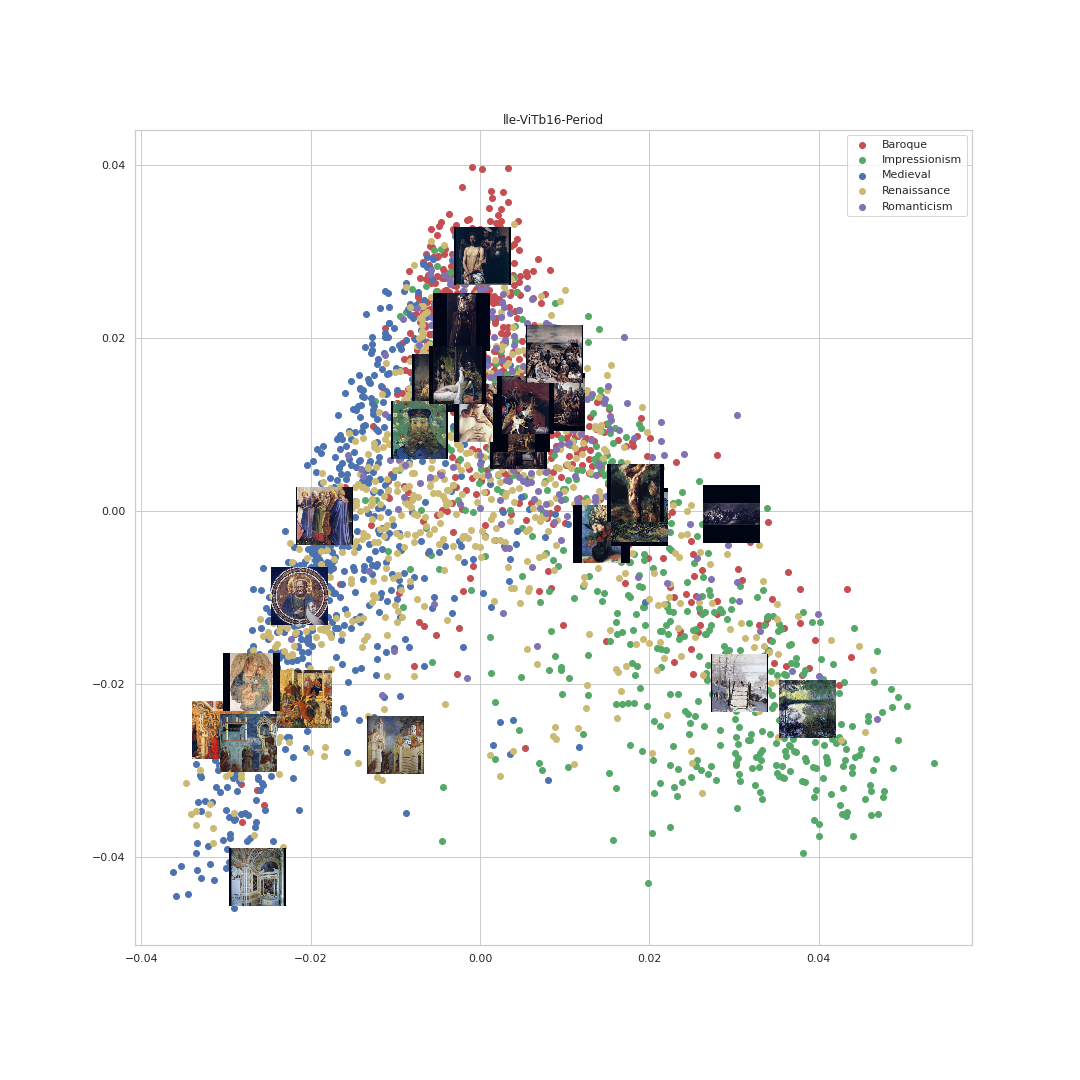}
    \caption{vit16-lle-period}
    \label{fig:vit16-lle-period}
\end{figure}

\begin{figure}[!ht]
    \centering
    \includegraphics[width=0.9\textwidth]{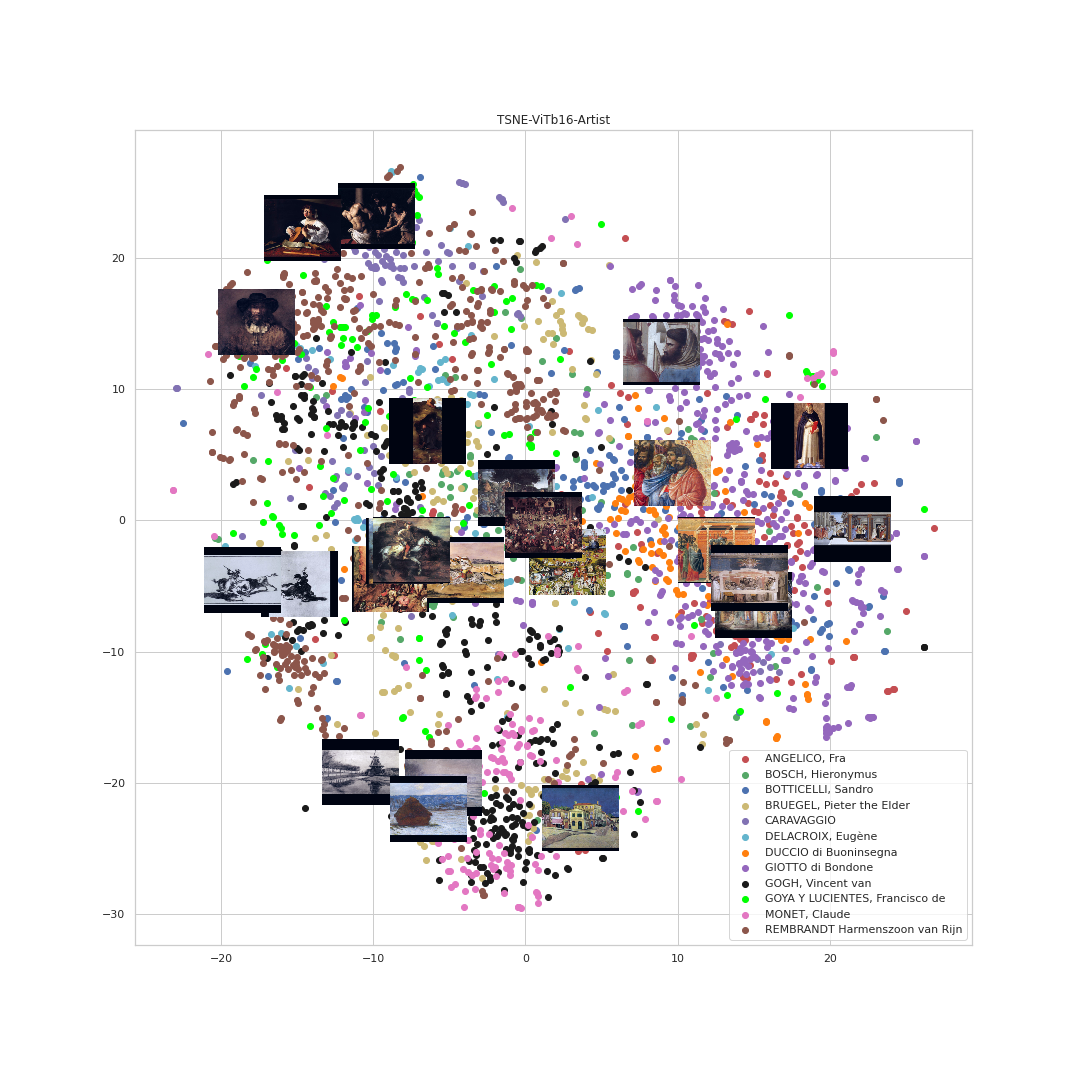}
    \caption{vit16-tSNE-artist}
    \label{fig:vit16-tsne-artist}
\end{figure}

\begin{figure}[!ht]
    \centering
    \includegraphics[width=0.9\textwidth]{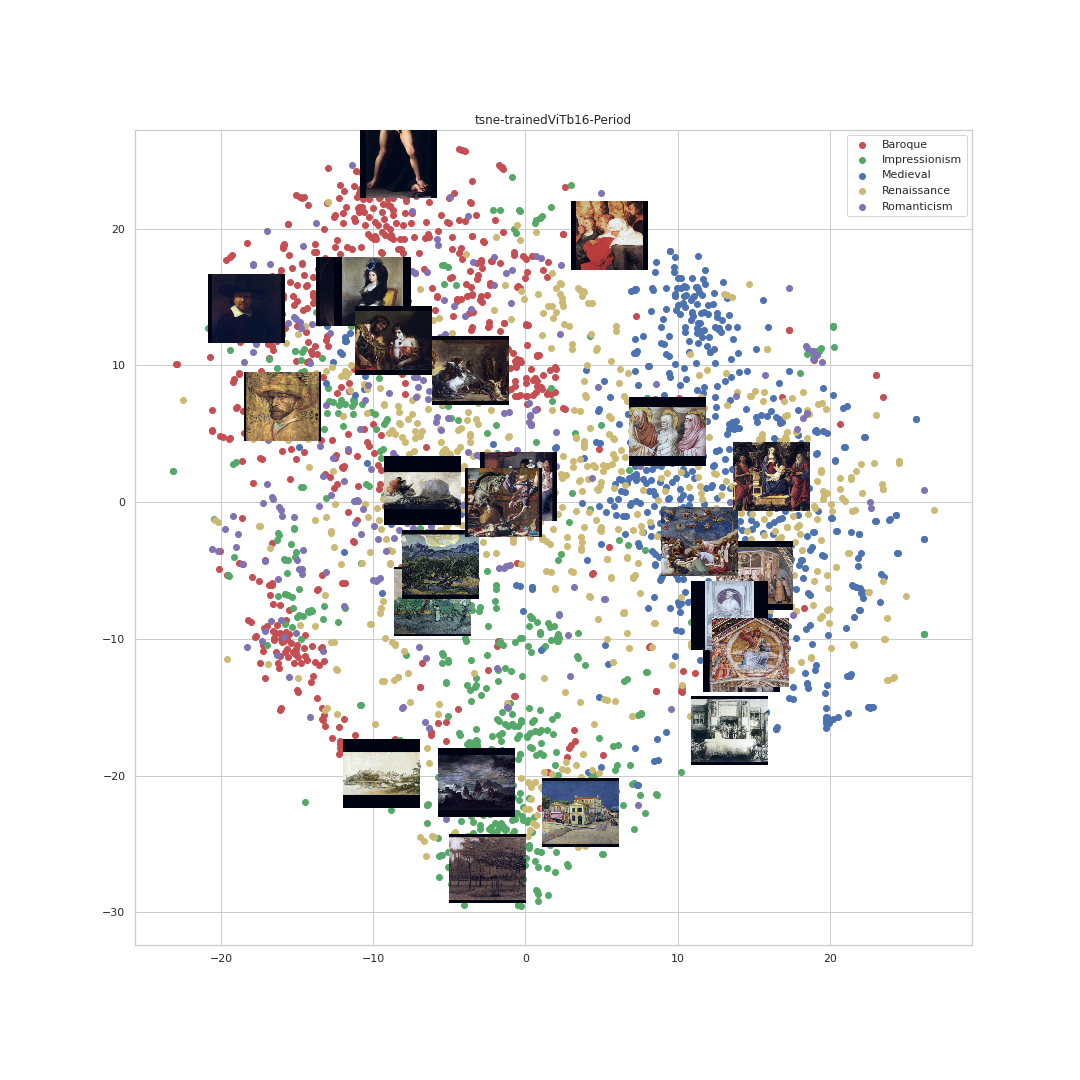}
    \caption{vit16-tSNE-period}
    \label{fig:vit16-tsne-period}
\end{figure}
       
\begin{figure}[!ht]
    \centering
    \includegraphics[width=0.9\textwidth]{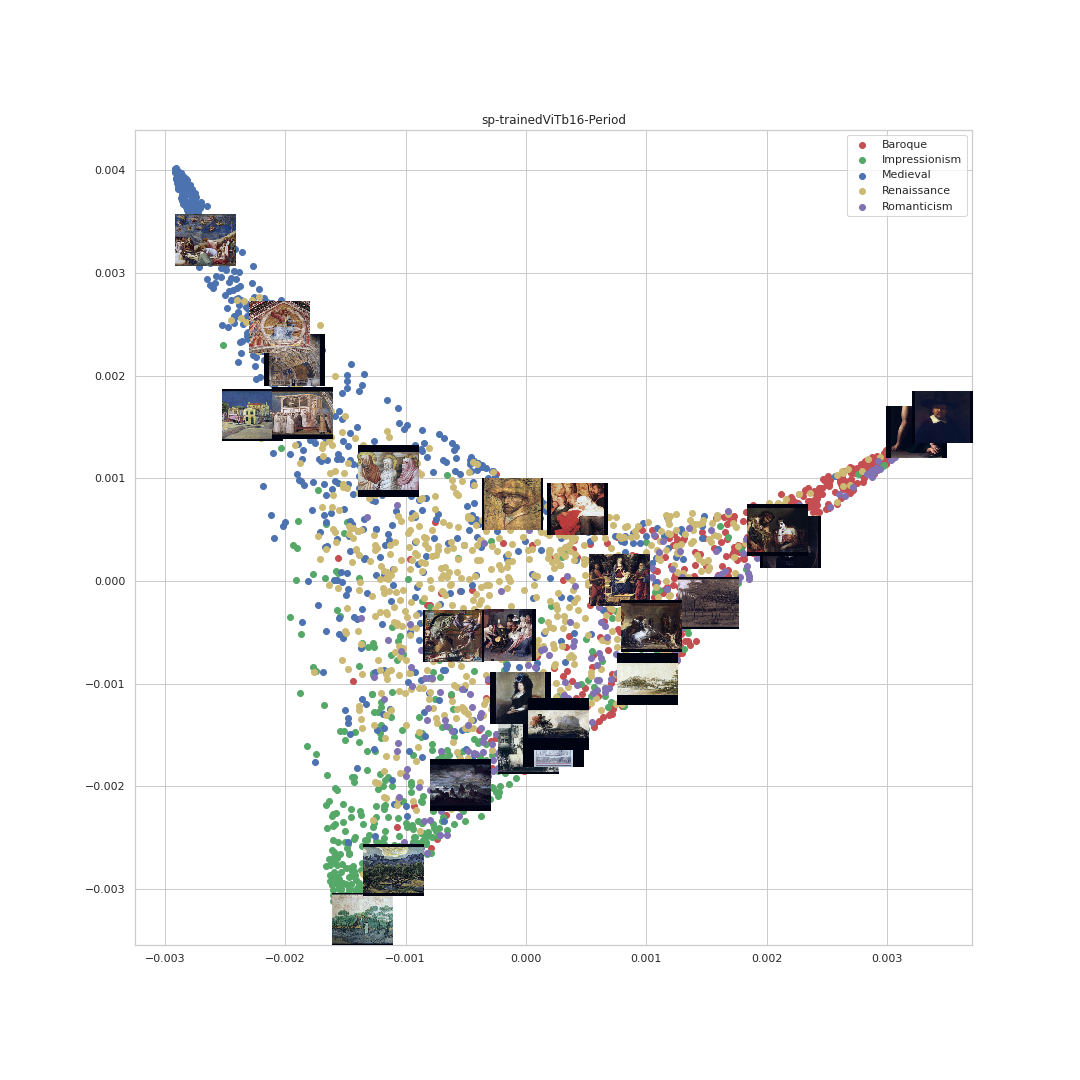}
    \caption{trained-vit16-Laplacian-Eigenmap-period}
    \label{fig:sp-trained-vit16-period}
\end{figure}

\begin{figure}[!ht]
    \centering
    \includegraphics[width=0.9\textwidth]{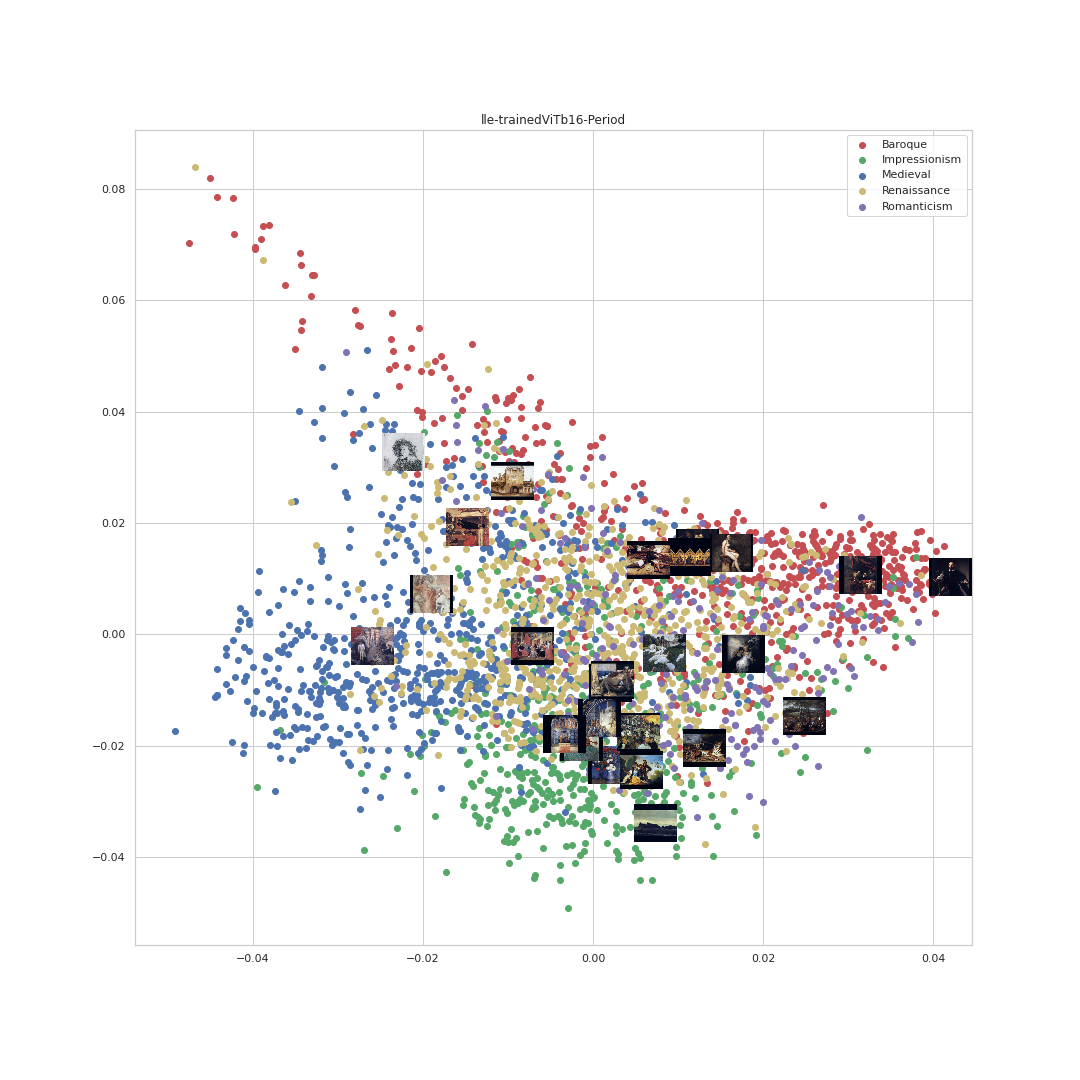}
    \caption{trained-vit16-lle-period}
    \label{fig:lle-trained-vit16-period}
\end{figure}

\begin{figure}[!ht]
    \centering
    \includegraphics[width=0.9\textwidth]{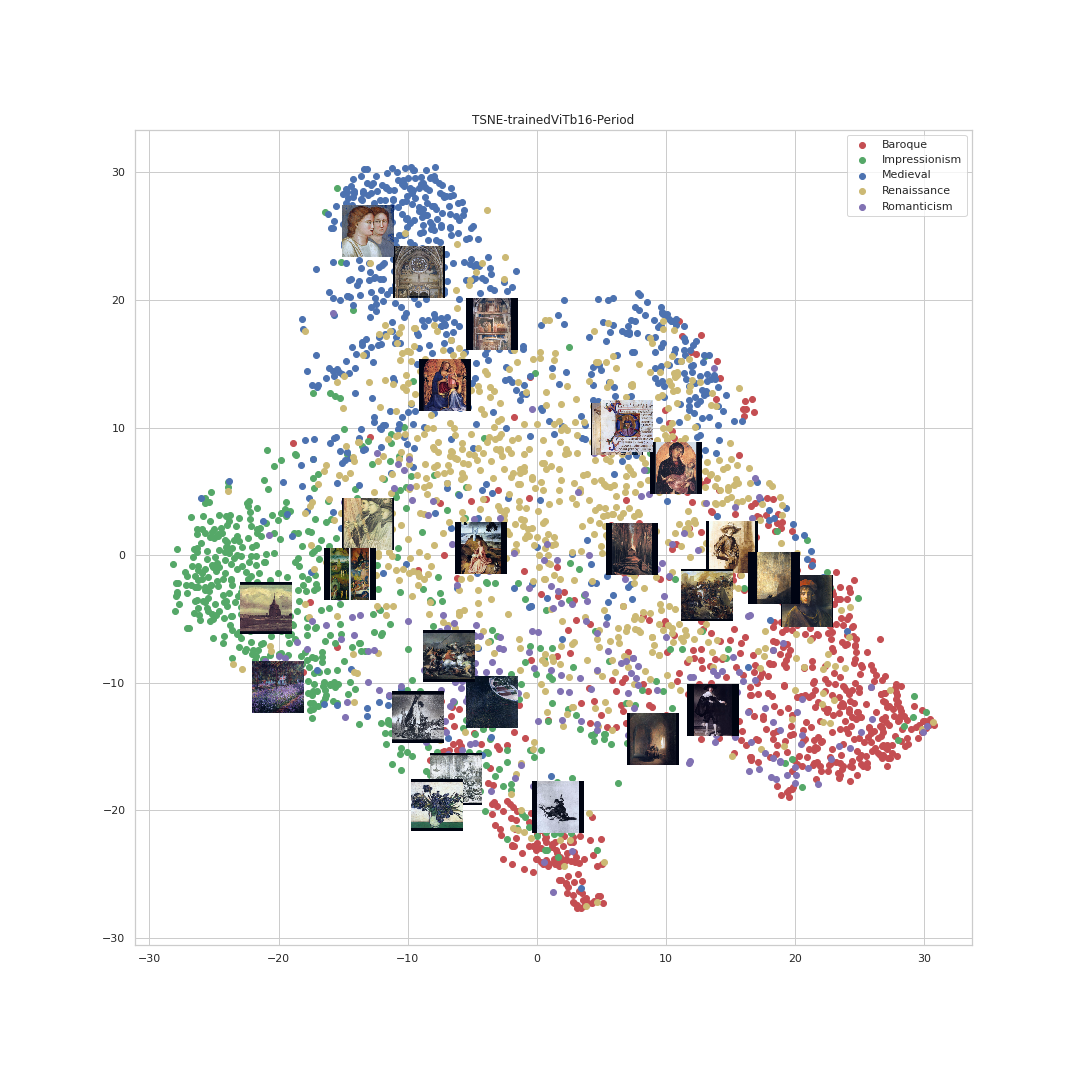}
    \caption{trained-vit16-tSNE-period}
    \label{fig:tsne-trained-vit16-period}
\end{figure}

\begin{figure}[!ht]
    \centering
    \includegraphics[width=0.9\textwidth]{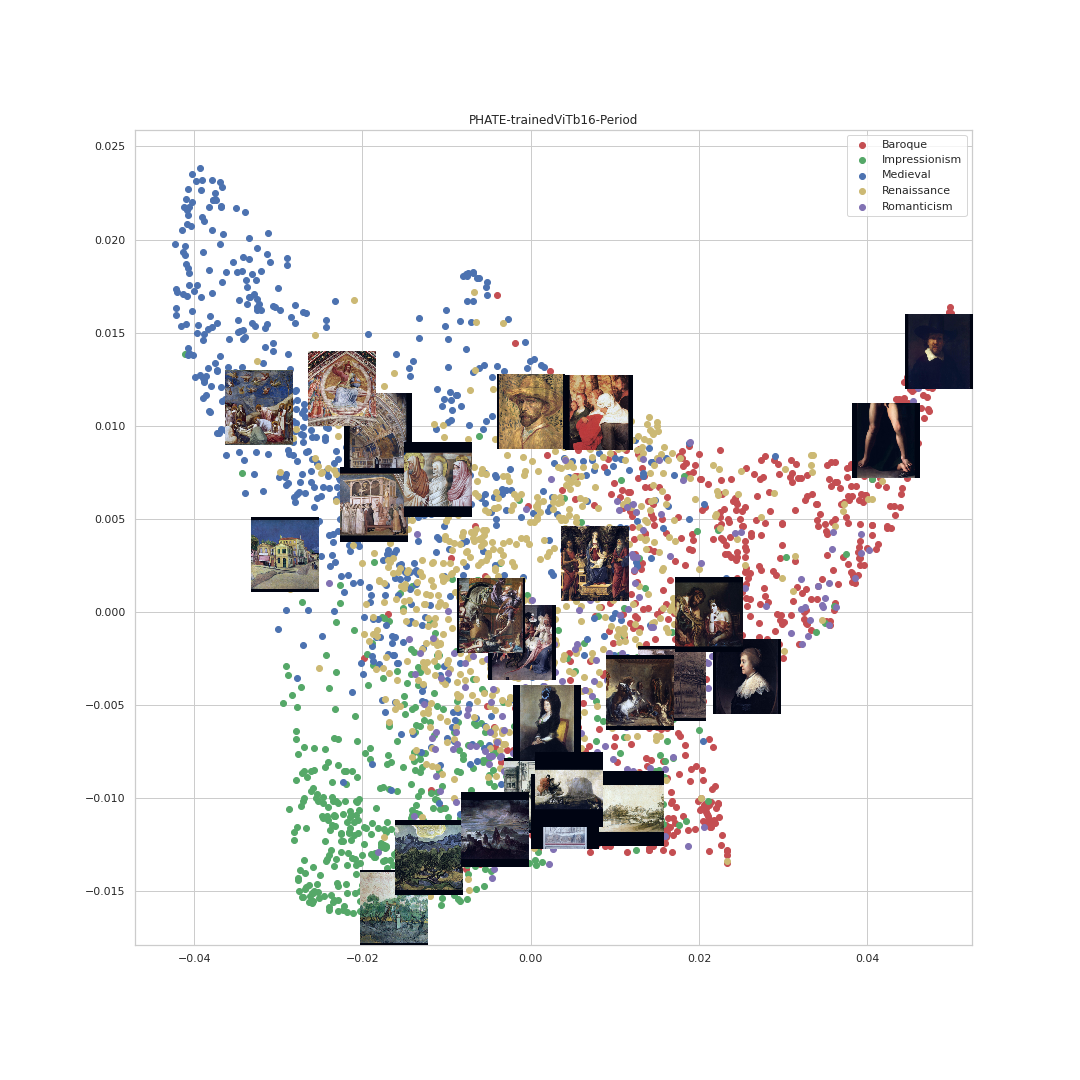}
    \caption{PHATE-trained-vit16-tSNE-period}
    \label{fig:phate-trained-vit16-period}
\end{figure}

\begin{figure}[!ht]
    \centering
    \includegraphics[width=0.9\textwidth]{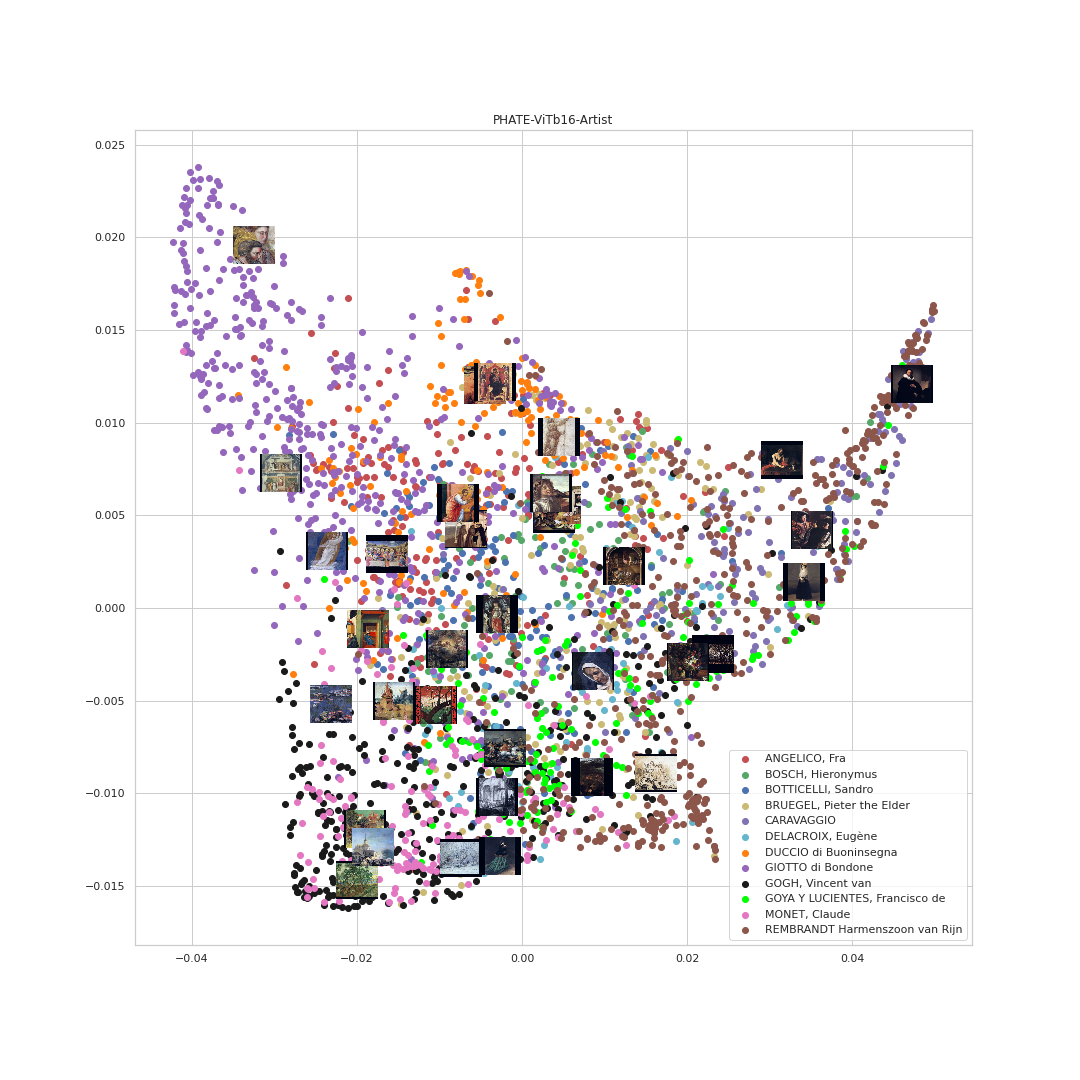}
    \caption{PHATE-trained-vit16-tSNE-artist}
    \label{fig:phate-trained-vit16-artist}
\end{figure}

\begin{figure}[!ht]
    \centering
    \includegraphics[width=0.9\textwidth]{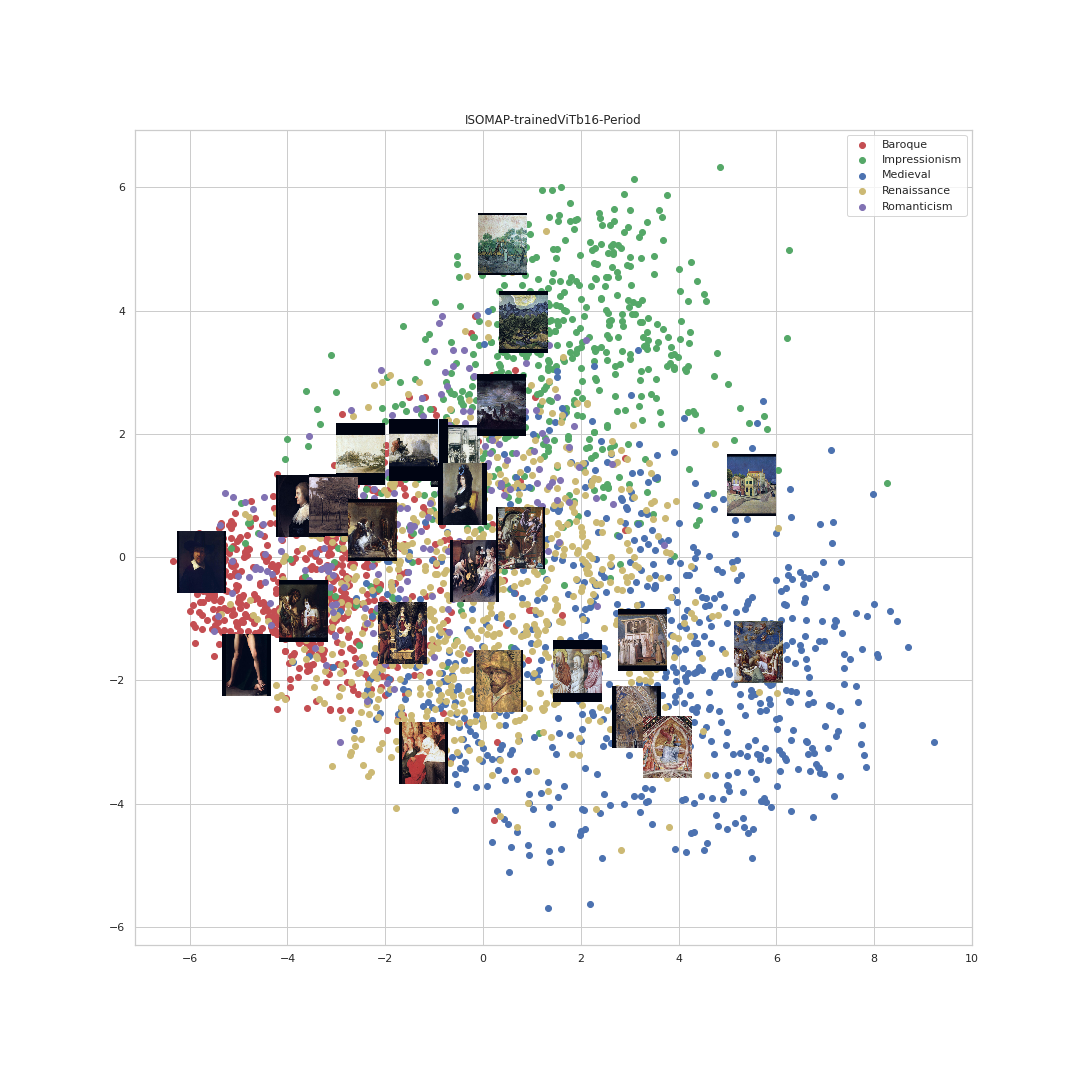}
    \caption{trained-vit16-ISOMAP-period}
    \label{fig:ISOMAP-trained-vit16-period}
\end{figure}

\end{document}